\documentclass[Afour,sageh,times,doublespace]{sagej}
\usepackage{amsmath,amssymb}
\usepackage{booktabs}
\usepackage{graphicx}
\usepackage{url}
\newcommand{\task}[1]{\textit{#1}}
\renewcommand{\journalname}{The International Journal of Robotics Research}

\newenvironment{orcidids}%
{\subsection*{\normalsize\sagesf\bfseries ORCID iDs}\begin{refsize}\noindent\raggedright\urlstyle{same}}%
{\par\end{refsize}}

\begin{document}

\runninghead{Chang Nie et al.}

\title{Towards the Vision-Sound-Language-Action Paradigm: The HEAR Framework for Sound-Centric Manipulation}

\author{
    Chang Nie\affilnum{1},
    Tianchen Deng\affilnum{1},
    Guangming Wang\affilnum{2},
    Zhe Liu\affilnum{1} and
    Hesheng Wang\affilnum{1}
}

\affiliation{
    \affilnum{1}School of Automation and Intelligent Sensing, Shanghai Jiao Tong University, Shanghai Key Laboratory of Navigation and Location Based Services, Shanghai 200240, China.\\
    \affilnum{2}Department of Engineering, University of Cambridge, Cambridge, CB2 1TN, UK.
}

\corrauth{
    Hesheng Wang,
    School of Automation and Intelligent Sensing, Shanghai Jiao Tong University, Shanghai Key Laboratory of Navigation and Location Based Services, Shanghai 200240, China.
}

\email{wanghesheng@sjtu.edu.cn}

\begin{abstract}
Humans and animals use sound as a crucial cue for interacting with the physical world, as acoustic events can reveal contact, completion, hidden contents, or process state. Embodied agents should similarly benefit from auditory awareness during manipulation, yet existing Vision-Language-Action (VLA) policies typically rely on persistent visual observations, while audio-aware variants often treat audio as speech, waveform renderings, or fixed pre-execution context. Such interfaces can miss transient sounds, such as beeps, clicks, rattles, or collision cues, especially under system latency and open-loop action chunking. We formalize this timing failure as the Blind Execution Interval (BEI), in which critical acoustic evidence may occur after an action chunk begins but disappear before the next policy update. To address this challenge, we introduce Vision-Sound-Language-Action (VSLA), a continuous control paradigm conditioned on vision, streaming audio, language, and proprioception under delayed decision loops. We further present HEAR, a VSLA framework that preserves causal auditory context across execution gaps, performs multimodal reasoning, models near-future audio dynamics during training, and generates smooth action chunks for closed-loop manipulation. To support learning and evaluation, we introduce OpenX-Sound for robotics-specific audio-visual-action pretraining and HEAR-Bench, a benchmark for sound-centric manipulation with strict causal timing constraints. On HEAR-Bench, HEAR achieves an 81\% success rate, outperforming waveform, ASR, and compact audio-native baselines, and reaches 70\% sound-causal success across four real-world Franka tasks. These results show that robust sound-centric manipulation requires not only native audio input, but also causal auditory persistence and explicit temporal grounding. Code and videos are available at \url{https://hear.irmv.top}.
\end{abstract}


\maketitle

\section{Introduction}
\label{intro}

In sound-centric manipulation, physical interaction requires more than recognizing where objects are. A robot must also know when contact occurs, when a process changes state, and when an external event demands a response. Vision is the dominant sensing channel in modern robot learning because it provides rich spatial geometry and has benefited from strong visual representations \citep{deng2024compact, deng2025best3dscenerepresentation}. Yet many state changes are expressed most directly through sound: a click reveals contact, a rattle exposes hidden contents, a beep marks completion, and a sputtering process signals that an unseen transition has occurred. Auditory perception therefore complements vision as a practical sensing modality for physical interaction \citep{stein1993merging}.

Current Vision-Language-Action (VLA) models, including Octo \citep{octo}, OpenVLA \citep{openvla}, and $\pi_{0.5}$ \citep{pi0}, are effective when visual observations and language instructions contain the necessary task state. However, these systems still operate largely in a visually persistent world. Recent audio-aware approaches reduce this gap by transcribing speech with ASR \citep{whisper}, rendering short waveform segments as images \citep{ast}, or using audio for instruction understanding, intent inference, and contact perception \citep{guo2025omnivla,wei2025audio}. These are important steps, but they often treat audio as an episodic context signal: sound is heard before acting, converted into text, or summarized in a fixed window aligned with the visual observation. Such interfaces are useful for high-level reasoning, but they do not fully support the lower-level requirement of listening while acting.

\begin{figure}[t]
	    \centering
	    \includegraphics[width=\linewidth]{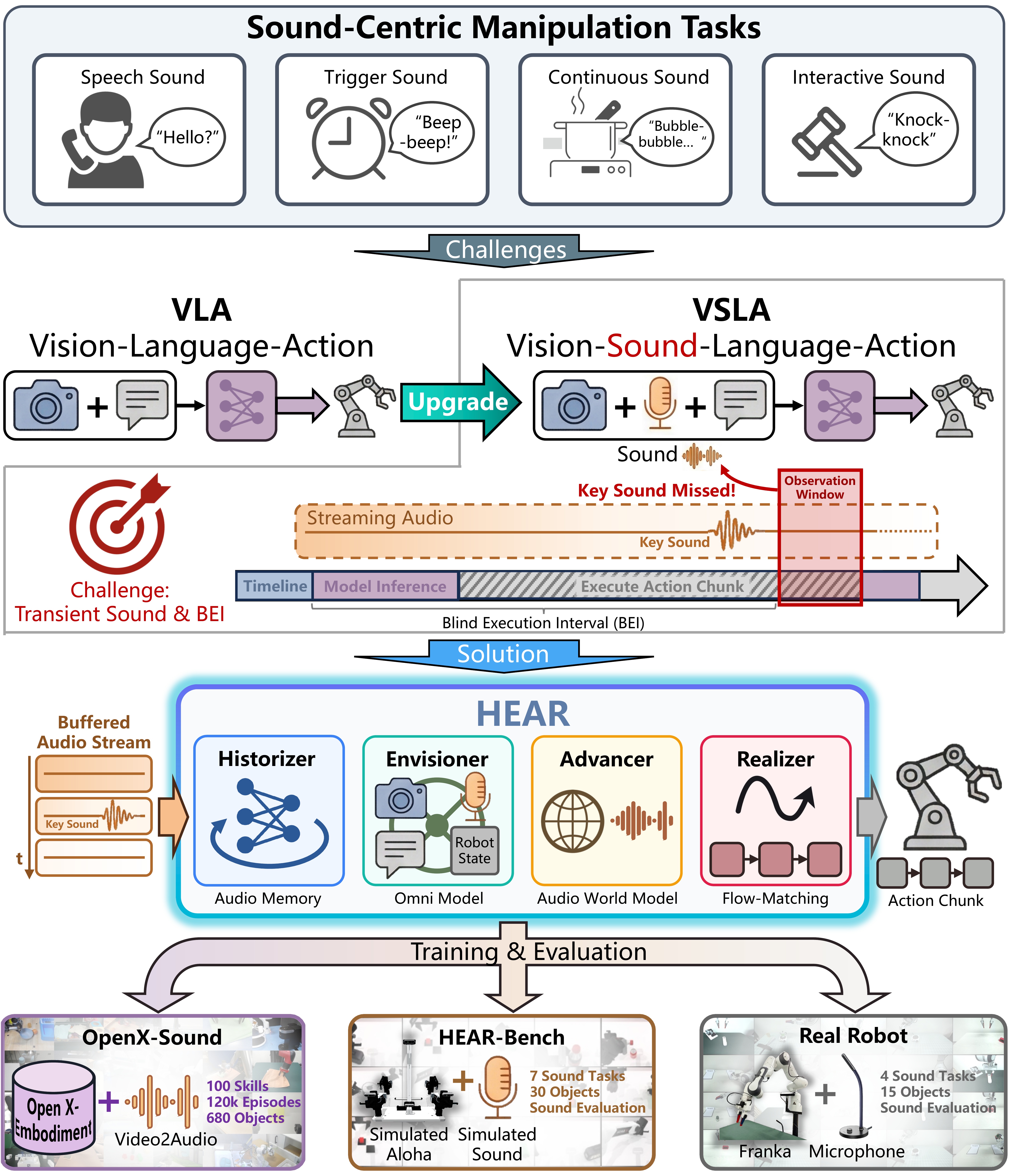}
\caption{\textbf{The HEAR framework for Vision-Sound-Language-Action (VSLA) manipulation.}
\textbf{Scenarios:} Sound-centric manipulation requires policies to use speech, trigger sounds, continuous process sounds, and interaction-generated sounds.
\textbf{Challenge:} Extending VLA policies to VSLA introduces a timing bottleneck: transient acoustic cues can fall inside the Blind Execution Interval (BEI) created by system latency and open-loop action chunking.
\textbf{Solution:} HEAR addresses this setting through causal audio memory (\textit{Historizer}), omni-sensory reasoning (\textit{Envisioner}), predictive audio dynamics (\textit{Advancer}), and smooth flow-matching action generation (\textit{Realizer}).
\textbf{Training \& Evaluation:} OpenX-Sound supports scalable robotics-specific pretraining, while HEAR-Bench and physical robot deployments provide sound-causal evaluation.}
	    \label{fig:headpic}
\end{figure}

The difference matters because visual and auditory information are distributed differently over time. Everyday manipulation scenes usually change gradually: an object, gripper, or appliance remains visible for many frames, so a modest perception delay often makes the image slightly stale rather than useless. Audio has the opposite structure. During quiet periods it may carry little task information, while critical evidence is concentrated in short events such as a tap, click, beep, phone ring, boiling transition, or material-dependent rattle. In the main text, we focus on the central consequence: sound-centric manipulation is not only a problem of adding an audio input; it is a problem of preserving brief acoustic evidence until the policy can act on it.

We organize everyday manipulation sounds into four categories: speech, event-triggered sounds, continuous process sounds, and physical interaction feedback. Speech can provide task-level intent or dialogue state, while the other categories expose low-level physical state that may be difficult or impossible to infer visually. Contact acoustics, for example, can serve as a low-cost proxy for material or tactile information \citep{huang2025tactile, sinapov2009}. These cues are valuable precisely because they are temporally specific, but that specificity also makes them fragile. If a microwave ``ding'' or an impact ``tap'' occurs between policy queries, a memoryless audio interface may either miss it entirely or represent it as a tiny, timing-ambiguous fragment inside a longer waveform. The robot may then make a visually plausible but causally wrong decision, such as acting before a required sound has occurred or failing to react after it has passed.

This vulnerability is amplified by the execution pattern of modern large robot policies. Because expensive policies cannot run at the low-level control frequency, they often generate action chunks and execute them open-loop for smooth motion \citep{act}. During this period, newly arriving observations cannot change the commands already being sent to the robot. We call the resulting gap the \emph{Blind Execution Interval} (BEI). Re-querying more frequently can shorten the gap, but it cannot eliminate sensing, preprocessing, inference, communication, and actuation latency. Aggressive replanning can also introduce discontinuities and robot ego-noise, which are harmful precisely when the robot is trying to listen. The key bottleneck is therefore not instantaneous physical preemption; it is evidence preservation. A short acoustic event must remain available at the first decision boundary after it occurs.

A second challenge is temporal grounding. Even when the BEI is short enough for transient cues to be preserved, the policy must still interpret acoustic progression across many decision boundaries during visually quasi-static intervals. Many sound-centric tasks require the robot to wait through such an interval, maintain a stable listening posture, and change behavior when the acoustic state evolves. A single camera frame may look nearly identical before and after a relevant acoustic phase boundary. Without a representation of temporal progress, chunked policies can drift, freeze, or collapse into low-motion action sequences during long waiting phases, a failure pattern related to Temporal Motion Collapse \citep{act}. A useful sound-aware policy must therefore do two things at once: remember transient evidence that may no longer be audible, and maintain a temporally grounded state that supports stable long-horizon behavior. Supplemental Section~S1 expands this motivation and the associated timing challenges.

To address these bottlenecks, we formalize \textbf{Vision--Sound--Language--Action (VSLA)} as a continuous-control setting conditioned on vision, streaming audio, language, and proprioception under delayed decision loops. We instantiate VSLA with \textbf{HEAR} (Figure~\ref{fig:headpic}), which augments a multimodal backbone with dedicated causal auditory memory and temporally grounded action generation. The \textit{Historizer} maintains a compact causal memory updated from streaming audio packets, allowing evidence that arrives between policy queries to condition the next action. The \textit{Envisioner} fuses vision, language, robot state, the current audio window, and this causal memory. It builds on Qwen3-Omni's general multimodal pretraining; within HEAR, its role is to combine that capability with the Historizer state and proprioception for temporally grounded action generation. The \textit{Advancer} provides a training-time temporal-grounding objective by predicting near-future discrete audio tokens, echoing recent evidence that future audio prediction can support manipulation when visual observations are ambiguous~\citep{zhang2025audioworldmodels}, and is removed at deployment. The flow-matching \textit{Realizer} produces smooth joint-position chunks, reducing discontinuous motion and ego-noise that can mask acoustic cues during listening.

Progress in this direction also requires data and evaluation protocols that reflect the timing structure of sound. We introduce OpenX-Sound, which augments selected Open X-Embodiment episodes with temporally aligned audio tracks generated from video \citep{vatt,difffoley}. Because the base Qwen3-Omni backbone already has broad multimodal pretraining on general audio-video data, OpenX-Sound is used as a robotics-specific audio-visual-action temporal-alignment stage rather than as generic multimodal pretraining or a replacement for native robot audio. We also introduce HEAR-Bench, a real-time event-conditioned audio benchmark built on RoboTwin2. Unlike standard visual benchmarks, HEAR-Bench streams task-relevant sounds and enforces sound-causal success rules that reject visually plausible actions completed before the required cue; physical robot experiments provide initial real-acoustic validation beyond this event-conditioned simulation.

We evaluate HEAR in HEAR-Bench and on a physical Franka Panda. HEAR outperforms ASR, waveform-adapter, and compact audio-native baselines, with the largest gains on tasks where brief cues or long-horizon acoustic processes make fixed-window audio insufficient. The ablations and failure decompositions show that the Historizer is the main mechanism for preserving acoustic evidence across the BEI, while the Advancer and Realizer provide targeted temporal-grounding and smooth-action benefits. At the same time, the real-robot results reveal that long-horizon physical tasks remain system-level deployment challenges, motivating future work on acoustic sensing, source separation, recovery policies, and longer-term memory.

Our primary contributions are summarized as follows:
\begin{itemize}
\item \textbf{The VSLA paradigm and BEI formulation.} We formalize \emph{Vision--Sound--Language--Action} (\textbf{VSLA}) as a delayed, chunked control setting in which streaming audio, vision, language, and proprioception jointly condition action generation. We identify the \emph{Blind Execution Interval} (BEI) as a system-level timing bottleneck that can make transient acoustic evidence unavailable, weakly represented, or temporally ambiguous for memoryless policies.
\item \textbf{The HEAR framework.} We propose \textbf{HEAR}, a VSLA policy architecture that combines causal audio memory (\textit{Historizer}), multimodal reasoning (\textit{Envisioner}), training-time temporal grounding through predictive audio dynamics (\textit{Advancer}), and smooth flow-matching action generation (\textit{Realizer}).
\item \textbf{Robotics-specific data and sound-causal evaluation.} We introduce \textbf{OpenX-Sound} as a robotics-specific audio-visual-action temporal-alignment pretraining resource and \textbf{HEAR-Bench} for event-conditioned, sound-causal manipulation evaluation under strict timing rules. Across the evaluated HEAR-Bench and Franka tasks, HEAR improves over the evaluated baselines on sound-causal manipulation.
\end{itemize}
To keep the main manuscript focused on the VSLA/BEI formulation, the HEAR mechanism, and the main simulation and real-robot evidence, an expanded motivation discussion, lower-level dataset construction details, implementation settings, and extended auxiliary diagnostics are provided in the Supplemental Material.

\section{Related Work}

\subsection{Vision--Language--Action Policies}

Recent work in robot learning has established the effectiveness of VLA policies. Earlier systems frequently used large language models to plan high-level steps while leaving low-level control to separate modules~\citep{saycan, voxposer}. The current trend has moved toward end-to-end models that map language instructions and camera images directly to continuous robot actions~\citep{rt1, palme, nie2025ermv, rt2}. Supported by large datasets like Open X-Embodiment~\citep{openx}, models such as Octo~\citep{octo} and OpenVLA~\citep{openvla}, and recent 3D or dexterous foundation policies~\citep{yang2026fp3, zhang2026dexora} have achieved strong performance across different robot platforms. A common assumption in these architectures is that the dominant evidence is visual. Because visual scenes typically change smoothly and objects remain in the camera's view over time, these models can safely rely on discrete image snapshots taken at a relatively low frequency.

To manage the heavy computation of these large models and ensure smooth robot movement, most VLA systems incorporate action chunking~\citep{act}. Instead of predicting one step at a time, the policy generates a short sequence of actions and executes them without waiting for new sensor updates. This approach has been widely adopted and improved by generative models like Diffusion Policy~\citep{diffusion_policy} and flow-matching methods such as $\pi_{0.5}$~\citep{pi0, flow_matching}; recent analyses further show that the execution horizon of chunked VLA policies strongly affects the trade-off between adaptability and stable control~\citep{wang2026vlalimits}. While this open-loop execution effectively hides computation delay, it creates a temporary gap in perception, which we refer to as the Blind Execution Interval (BEI).

For ordinary visual tasks, the BEI is usually manageable because the visual state persists across the execution gap. However, this architectural design becomes a significant bottleneck when researchers attempt to incorporate highly transient signals like audio. A brief acoustic trigger, such as a click, can easily occur and finish while the robot is executing an action chunk. At the next policy query, the cue may be outside a short observation window, or it may survive only as a brief, timing-ambiguous component inside a longer mixed waveform. Therefore, extending current VLA models to sound-centric tasks requires new mechanisms to maintain continuous sensory memory across these execution gaps. Our work builds upon the continuous control capabilities of existing flow-matching models but specifically addresses this timing mismatch so fleeting sounds remain available as decision-relevant evidence.

\subsection{Audio in Robotics}

Audio has historically played a variety of roles in robotics, ranging from high-level human interaction to low-level physical feedback. For high-level tasks, audio has traditionally been treated as a communication channel. Many systems utilize automatic speech recognition (ASR) modules \citep{wav2vec2,whisper} to convert spoken commands into text prompts. Recent advancements have sought to integrate speech more directly into VLA models. For example, frameworks like SVA and VLAS explicitly process speech instructions to retain speaker identity and prosody, which are otherwise lost in pure text transcription \citep{li2025sva, zhao2025vlas}. While highly effective for human-robot interaction, these methods primarily focus on explicit verbal commands rather than the ambient acoustic physics generated during manipulation. In the domain of navigation, agents frequently use sound to localize sources or infer off-screen events \citep{soundspaces}. However, navigation agents can typically move their base to improve listening geometry over time, whereas manipulation requires reacting to brief temporal events exactly when the robot is generating significant mechanical ego-noise.

Beyond direct commands, recent works have explored audio for broader instruction understanding and proactive intent inference. For instance, audio-visual models can utilize sound to disambiguate visually identical objects when language commands are otherwise unclear \citep{guo2023audio}. Notably, RoboOmni \citep{wang2025roboomni} studies proactive manipulation within an omni-modal context, where a robot infers human intent by combining dialogue with environmental sounds. While these approaches represent a significant step toward multi-sensory intelligence, they primarily utilize audio to establish context or infer intent prior to, or around, the execution of an action. They generally do not treat audio as a low-level, real-time physical feedback signal during contact, nor do they address the temporal misalignment between transient acoustic events and the chunked, open-loop execution of modern VLA policies.

At the level of physical interaction, a substantial body of work leverages audio as a direct feedback mechanism. Early interactive perception studies demonstrated that robots could estimate material properties based on acoustic responses \citep{sinapov2009,sinapov2014}. Building on this, recent efforts have utilized sound to assess the stability and gentleness of robotic grasps \citep{nakahara2025learning}. To scale these capabilities, the community is rapidly moving toward multisensory foundation models that add contact audio or unify multiple sensor streams in shared representations \citep{guo2025omnivla, liu2025mla, wei2025audio}. Other research has successfully fused contact microphones with vision, demonstrating that audio can serve as a scalable proxy for tactile sensing \citep{seehearfeel,hearingtouch}. Complementary to passive listening, active acoustic sensing injects known signals to infer contact states from the acoustic response \citep{activeacoustic,wall2023acoustic} or uses active excitation through the gripper for long-horizon tasks \citep{zhang2025vibecheck}. To process this sensory data, existing methods often plot short audio clips into two-dimensional waveform images, effectively treating sound as an additional static picture, or they rely on fixed-length observation windows strictly aligned with the camera frame rate.

These models have substantially improved multimodal representation learning, but they still face structural challenges when deployed in continuous, real-time control loops. Processing static audio clips or text transcripts works well for identifying loud events or receiving initial instructions. However, it does not naturally support continuous listening during execution. This structural challenge is exacerbated by the modern control paradigm itself. Because large robot policies often execute actions in open-loop chunks to maintain smooth motion, they create temporary gaps in perception. If a brief acoustic event, such as a microwave ``ding'', occurs during an execution gap, a static query-time representation may either miss it under short-window or adverse-alignment settings, or blur its timing and causal role inside a longer observation window.

Scaling auditory capabilities often involves leveraging large-scale pre-trained audio representations, such as CLAP \citep{clap}, ImageBind \citep{imagebind}, and BEATs \citep{beats}. However, these foundation models are generally designed to process fixed-length audio clips offline and independently. When deployed in a real-time robotic system, they operate statelessly. This means they do not inherently maintain a temporal context from one clip to the next. Adapting powerful semantic foundation models to operate causally across these discrete control steps remains difficult. Rather than attempting to process a literal continuous analog stream, a primary motivation of our work is to build a general paradigm that maintains temporal context from one clip to the next. By establishing a causal memory across discrete audio windows, we ensure that transient acoustic signals occurring between policy queries are preserved and systematically synchronized with the robot's control loop.

\subsection{Datasets and Benchmarks}
The scaling of VLA policies has been enabled by standardized robot datasets. Open X-Embodiment~\citep{openx}, for example, aggregates visual-kinematic trajectories across many tasks and embodiments, but does not include synchronized microphone recordings. Existing audio-rich manipulation datasets demonstrate the value of acoustic feedback, often using specialized ear-in-hand microphones or contact sensors during carefully controlled collection \citep{du2022playitbyear,hearingtouch}. These datasets are valuable, but their hardware requirements and acoustic sensitivity make them difficult to scale across laboratories. OpenX-Sound addresses this pretraining bottleneck by retrospectively adding temporally aligned audio tracks to selected Open X-Embodiment episodes using video-to-audio generation models \citep{vatt,difffoley}. This synthetic audio is used as a scalable temporal-alignment signal for robotics pretraining, consistent with recent work on generative audio for multimodal sim-to-real robot policies \citep{wang2025sound}; it is not treated as a substitute for native robot audio during final deployment.

Evaluation presents a separate bottleneck. Standard manipulation benchmarks such as RLBench \citep{rlbench} and ManiSkill \citep{maniskill2} focus on spatial reasoning and visual control in tasks without task-relevant audio. Simulation platforms with acoustics, such as SoundSpaces \citep{soundspaces}, primarily target navigation toward continuous sound sources rather than sound-causal manipulation. HEAR-Bench fills this evaluation gap with event-conditioned audio playback, streaming microphone observations, and success rules that reject premature visually plausible actions. This lets us test whether a policy can preserve and act on acoustic evidence under delayed, chunked execution rather than merely reaching a final visual state.

\section{The VSLA Paradigm}
\label{sec:formulation}
We formalize the Vision-Sound-Language-Action (VSLA) paradigm. Standard Vision-Language-Action (VLA) implementations predominantly rely on visual and proprioceptive observations to infer environmental states. In contrast, the VSLA paradigm focuses on sound-centric manipulation tasks, where critical physical states or event timings are actively conveyed through acoustic cues. In these scenarios, integrating continuous audio is essential, as sound provides low-level physical feedback and precise success verification that cameras and joint encoders alone cannot capture.

The core theoretical challenge in VSLA is bridging the fundamental temporal mismatch between the continuous, high-frequency nature of acoustic events and the delayed, discrete update cycles of modern robotic policies. To systematically address this, the following subsections define the VSLA observation interface, model the structural blind spots induced by chunked execution, and formalize the timing-sensitive success criteria required for evaluation.

\subsection{VSLA Observation and Action Space}

\subsubsection{Environment and Signals}
We consider a dynamical system evolving in continuous physical time $\tau \in \mathbb{R}_{\ge 0}$. The robot runs a low-level controller at frequency $f_c$ Hz, which induces discrete control cycles indexed by $t \in \mathbb{N}$. The physical time corresponding to cycle $t$ is $\tau_t \triangleq t / f_c$.

At each control cycle $t$, the robot may receive observations from multiple modalities:
\begin{itemize}
    \item \textbf{Vision} Multi-view RGB images $I_t^{1:V}=\{I_t^{(1)}, \dots, I_t^{(V)}\}$.
    \item \textbf{Proprioception} The robot kinematic state $s_t \in \mathbb{R}^{d_{\text{qpos}}}$, consisting of joint positions and gripper degree(s) of freedom.
    \item \textbf{Instruction} A natural-language instruction $l$, fixed within an episode.
    \item \textbf{Streaming audio} A microphone samples a continuous waveform at rate $f_s$ Hz. Let $\mathcal{A}_{n} \in \mathbb{R}$ denote the $n$-th audio sample.
\end{itemize}

To connect continuous-time audio to discrete control cycles, define the most recent audio sample index available at physical time $\tau$ as $n(\tau) \triangleq \lfloor f_s \tau \rfloor$. At control cycle $t$, the most recent available sample index is
\begin{equation}
n_t \triangleq n(\tau_t) \;=\; \left\lfloor t \cdot \frac{f_s}{f_c} \right\rfloor .
\label{eq:fc_fs_mapping}
\end{equation}
We denote the audio history up to cycle $t$ as $\mathcal{H}^{\text{audio}}_{t} \triangleq (\mathcal{A}_{0}, \dots, \mathcal{A}_{n_t})$.

\subsubsection{Observation Interface}
\label{sec:vsla_obs}

The policy is queried only at a subset of control cycles, denoted by decision times $\{t_k\}_{k\ge 0}$. Real systems also exhibit non-zero end-to-end latency. We model this latency as an observation delay of $\tau_{\text{sys}}$ control cycles, meaning that at decision time $t_k$ the freshest sensor data corresponds to the delayed index
\begin{equation}
\bar{t}_k \triangleq t_k - \tau_{\text{sys}} .
\label{eq:delayed_time}
\end{equation}
We treat $\tau_{\text{sys}}$ as an integer number of control cycles. If $t_k < \tau_{\text{sys}}$, we clip $\bar{t}_k$ to 0.

For audio, the interface provides a short, causal segment rather than the entire past waveform. Intuitively, at decision time $t_k$ the policy receives the most recent $T_{\text{win}}$ seconds of audio that have arrived by the delayed time $\bar{t}_k$. Let $N_{\text{win}} \triangleq \lfloor f_s T_{\text{win}} \rfloor$. The window is defined as the last $N_{\text{win}}$ samples up to index $n_{\bar{t}_k}$, padded with zeros when the episode starts:
\begin{equation}
\begin{split}
A_{t_k}[i] &\triangleq
\begin{cases}
\mathcal{A}_{n_{\bar{t}_k}-N_{\text{win}}+1+i}, & n_{\bar{t}_k}-N_{\text{win}}+1+i \ge 0,\\
0, & \text{otherwise},
\end{cases} \\
&\quad \text{for } i=0,\dots,N_{\text{win}}-1 .
\end{split}
\label{eq:audio_window}
\end{equation}
The per-decision VSLA observation is then
\begin{equation}
o_{t_k} \triangleq \big(I_{\bar{t}_k}^{1:V},\, A_{t_k},\, l,\, s_{\bar{t}_k}\big).
\label{eq:obs}
\end{equation}
This interface makes the delayed decision loop explicit while keeping the audio stream causal.

\subsubsection{Action Space}
The robot is controlled by a command vector $a_t \in \mathbb{R}^{d_{\text{qpos}}}$, where $d_{\text{qpos}}$ is the embodiment-specific joint-position dimension including gripper degree(s) of freedom. A low-level controller tracks these commands to drive the physical system.

\subsection{Temporal Mismatch and the Blind Execution Interval}

\subsubsection{Persistence Mismatch Between Vision and Sound}
\label{sec:vsla_timing}
A fundamental distinction between vision and sound in typical manipulation tasks is their behavior under temporal delay. Visual observations primarily encode spatial geometry. Because physical objects possess inertia and robots operate under kinematic constraints, visual states change continuously and remain informative across multiple control cycles. If a camera drops a frame, the object's state can often be reliably interpolated.

In contrast, decision-critical acoustic cues are essentially discrete physical events. They represent abrupt energy releases, such as an impact click or an electronic beep, making them highly transient and non-repeatable. The steady-state audio between these events typically contains low-information ambient noise. When the decision loop is delayed or operates at a low rate, these high-density transient cues may easily fall between observation windows or occupy only a tiny fraction of a later waveform query. Unlike visual states, a poorly retained acoustic impulse can leave the policy without a reliable representation of the state transition that produced it.

\subsubsection{Decision Cadence and Latency}
\label{sec:decision_cadence}
Let $\{t_k\}_{k\ge 0}$ denote the control cycles at which the policy is queried. We define the decision interval and an effective perception-action gap, measured in control cycles, as
\begin{equation}
\Delta_k \triangleq t_{k+1}-t_k,\qquad \tau_{\text{sys}} \ge 0,\qquad G_k \triangleq \Delta_k+\tau_{\text{sys}} .
\label{eq:decision_latency}
\end{equation}
Here $\Delta_k$ captures the update cadence of the policy and $\tau_{\text{sys}}$ aggregates sensing, preprocessing, communication, inference, and actuation delays. The quantity $G_k$ summarizes how long a transient event may need to persist in order to influence the next decision when only windowed observations are available.

\subsubsection{Action Chunking and the Blind Execution Interval}
\label{sec:async_inference}
Because large backbones are often queried at a rate below the low-level controller, modern systems frequently use action chunking. Let $t_k$ denote the control cycle at which the $k$-th inference is triggered. We denote an action chunk of length $H$ by
\begin{equation}
\mathbf{a}_{k} \triangleq (a_{t_k}, a_{t_k+1}, \dots, a_{t_k+H-1}).
\label{eq:chunking}
\end{equation}
At each decision, the policy predicts a chunk $\hat{\mathbf{a}}_k$ that is executed open-loop~\citep{act}. We denote the executed horizon by $H_{\text{exec}}$ control cycles, so the next query occurs at $t_{k+1} = t_k + H_{\text{exec}}$, which implies $\Delta_k = H_{\text{exec}}$. In the common case, $H_{\text{exec}}=H$ and the entire chunk is executed.

During the interval $(t_k, t_{k+1})$, new observations cannot affect the commands currently being executed. We refer to this execution gap as the \emph{Blind Execution Interval} (BEI).

\subsubsection{Sound Causality and Evidence Vanishing}
Transient sound combined with non-zero latency $\tau_{\text{sys}}$ and delayed or low-rate updates creates an important limitation in sound-centric manipulation. A brief cue can occur and end while the system is executing an open-loop chunk. If the next decision only receives a finite window of recent samples, the cue may either fall outside that window or remain only as a short, low-duty-cycle event mixed with several seconds of unrelated audio. In both cases, a query-time window alone provides a weak and timing-ambiguous representation of the causal event.

To make this concrete, consider the audio-sample indices used to form the next observation. Let a critical event occupy a contiguous sample interval $[n_e, n_e+D_e)$ in the microphone stream, where $n_e$ is the start sample index and $D_e$ is the duration in samples. At decision time $t_{k+1}$, the policy forms its audio input from the last $N_{\text{win}}$ samples available at the delayed time $\bar{t}_{k+1}$. This window covers the index interval $[n_{\bar{t}_{k+1}}-N_{\text{win}}+1,\, n_{\bar{t}_{k+1}}+1)$.

The strict dropout case occurs when the event ends before this window begins:
\begin{equation}
\begin{split}
n_e + D_e \le n_{\bar{t}_{k+1}}& - N_{\text{win}} + 1
\quad \Longrightarrow \\
[n_e,\, n_e+D_e) &\cap [n_{\bar{t}_{k+1}}-N_{\text{win}}+1,\, n_{\bar{t}_{k+1}}+1) = \emptyset .
\end{split}
\label{eq:evidence_vanishing}
\end{equation}
In other words, the acoustic event terminates strictly before the new observation window begins, so a window-only interface provides no direct evidence that the event occurred. This is the most severe case. In our deployed setting the nominal waveform window is longer than the measured BEI, so many cues are not literally absent from the next raw waveform. The remaining problem is representational: a 50--200\,ms event can be embedded in a 4.0\,s query-time window after the robot has continued executing, causing fixed-window encoders to blur event timing, phase, and causal relevance. The Historizer is therefore not introduced merely to extend the raw waveform duration; it maintains an explicit causal state over streaming packets so that short acoustic cues occurring during $(t_{k-1}, t_k)$ remain represented as decision-relevant evidence at the next policy query.

Concretely, we require a causal state update function $f_\varphi$ and a chunk policy $\pi_\theta$ such that:
\begin{equation}
\begin{aligned}
h_{t_k} &= f_\varphi\!\left(h_{t_{k-1}},\, \mathcal{A}_{n_{\bar{t}_{k-1}}+1 \,:\, n_{\bar{t}_k}}\right),\\
\hat{\mathbf{a}}_k &= \pi_\theta\!\left(o_{t_k},\, h_{t_k}\right).
\end{aligned}
\label{eq:vsla_policy_history}
\end{equation}
This requirement captures the need for sound-centric manipulation under delayed, chunked execution. Transient cues that occur during an execution gap must be summarized into $h_{t_k}$ so they can inform the next decision.

\subsection{Temporal Grounding and Motion Collapse}

Capturing a transient cue is necessary but not sufficient. Sound-centric tasks often include long monitoring phases in which the robot must hold a stable posture and wait, followed by a rapid reaction once an acoustic trigger occurs. During such phases, the visual scene can be quasi-static and provides weak information about temporal progression. Under these conditions, successive observations can be nearly indistinguishable for many decisions, even though the desired control behavior changes over time.

This creates temporal aliasing that is common in partially observable settings~\citep{heess2015memory}. There may exist two decision indices $k \neq k'$ such that the available observations are similar while the expert action chunks differ:
\begin{equation}
o_{t_k} \approx o_{t_{k'}} \quad \text{but} \quad \mathbf{a}_k^* \neq \mathbf{a}_{k'}^* .
\label{eq:temporal_aliasing}
\end{equation}
In intuitive terms, the robot experiences a ``waiting dilemma'': it sees the same static scene frame after frame, making it difficult to maintain a steady posture without an internal sense of elapsing time. In practice, this mismatch can lead to unstable waiting behavior under chunked execution, including drift or freezing. It relates closely to the Temporal Motion Collapse phenomenon reported for long-horizon chunked policies~\citep{act}.

A practical way to mitigate this issue is to encourage the policy representation to encode the flow of time even when vision changes little. Audio is well-suited for this role because it evolves continuously and often reflects ongoing physical processes. In our setting, this motivates auxiliary predictive objectives that use near-future audio as supervision, so that the latent representation learns to anticipate how the soundscape will evolve. This provides temporal grounding without introducing additional evaluation criteria, and it matches the constraints of streaming, causal deployment~\citep{arandjelovic2017look, jaderberg2016reinforcement}.

\subsection{Learning Objective and Causal Evaluation}

\subsubsection{Timed Success Criteria}
\label{sec:timed_success}
Sound-centric manipulation is timing-sensitive. Reaching the geometric goal before the relevant acoustic cue should be counted as failure, even if the final state looks visually correct.

Let $g(\mathcal{H}^{\text{audio}}_t)\in\{0,1\}$ be a task-specific predicate that becomes true when the cue condition is satisfied, for example when a beep starts. Define
$t_{\text{snd}}=\min\{t \in \{0,\dots,T\}: g(\mathcal{H}^{\text{audio}}_t)=1\}$.
Let $x_t$ denote the environment state used to evaluate task completion, for example privileged simulator state in benchmark settings. Let $T$ denote the maximum discrete length of the episode. Define
$t_{\text{goal}}=\min\{t \in \{0,\dots,T\}: \mathrm{Goal}(x_t)=1\}$,
with the convention $\min \emptyset = +\infty$.
We evaluate a trajectory $\xi$ using the timing-sensitive success indicator
\begin{equation}
\mathrm{Success}(\xi)=\mathbb{1}\!\left[t_{\text{snd}}\le t_{\text{goal}}\le T\right].
\label{eq:timed_success}
\end{equation}
This rule enforces sound causality at evaluation time by rejecting early actions that reach the goal before the cue and by treating trajectories without a cue event as failure.

\subsubsection{Learning Objective}
We adopt imitation learning from expert demonstrations. At each decision time $t_k$, the expert executes an action chunk of length $H$,
\begin{equation}
\mathbf{a}_{k}^* \triangleq (a_{t_k}^*, \dots, a_{t_k+H-1}^*),
\label{eq:bc_target}
\end{equation}
and the learner predicts a chunk $\hat{\mathbf{a}}_k$ conditioned on the available observation $o_{t_k}$ and a causal memory $h_{t_k}$.

At the formulation level, we treat the learner as a conditional action-chunk model trained to match expert behavior under the VSLA interface and its causal constraints. The concrete instantiation can be a deterministic regressor or a conditional generative model, and it may include auxiliary objectives to encourage causal persistence and temporal grounding. The specific training losses used in our approach are described in Section~\ref{sec:method}.

\begin{figure*}[t]
    \centering
    \includegraphics[width=0.9\linewidth]{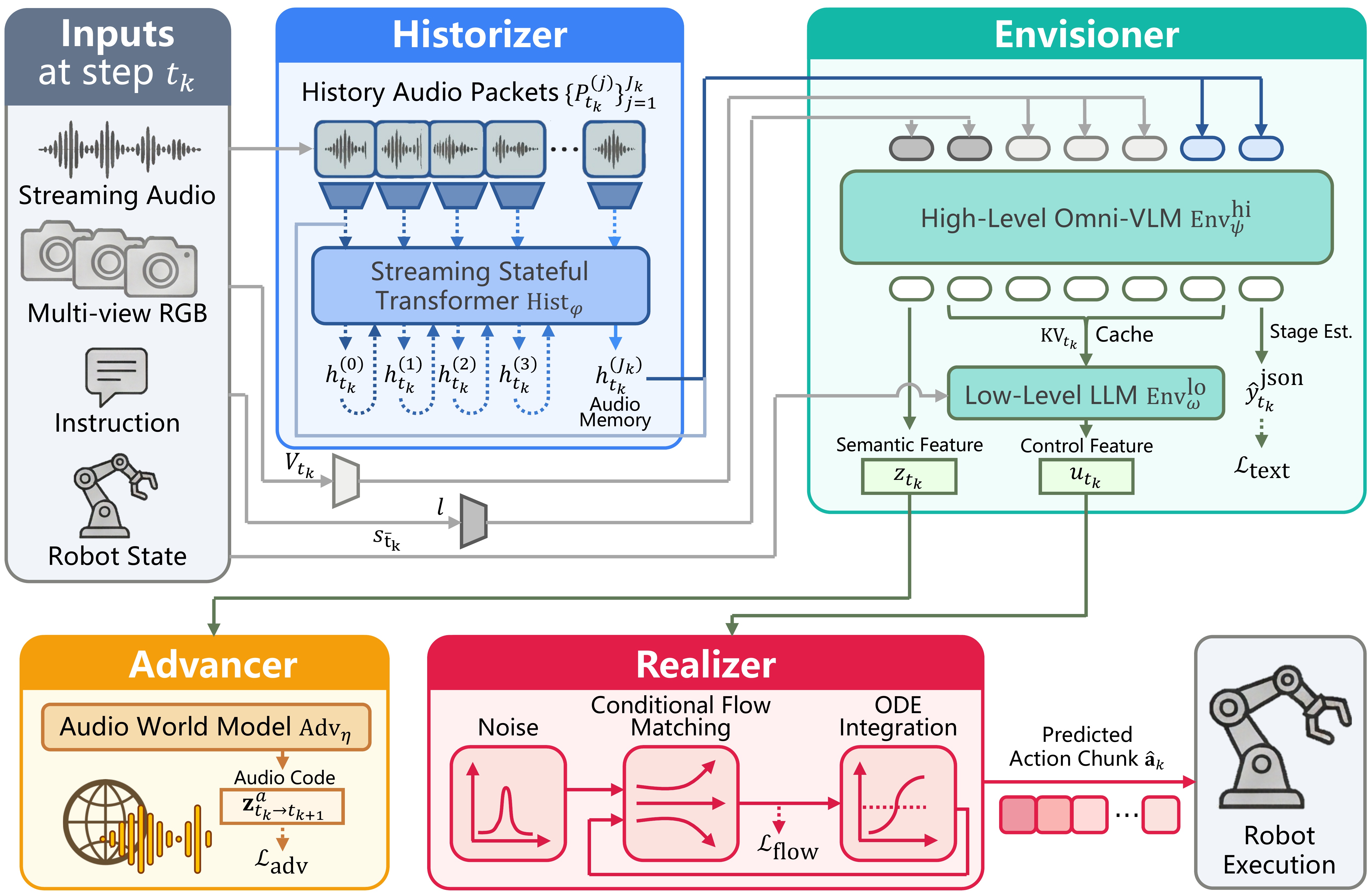}
    \caption{
    \textbf{The HEAR framework architecture.}
    The \textbf{Historizer} processes audio packets with a Streaming Stateful Transformer that maintains a compact causal memory $h_{t_k}$ and bridges execution gaps, including the Blind Execution Interval induced by open-loop chunking.
    The \textbf{Envisioner} employs a hierarchical design. Its high-level omni-modal model integrates multimodal inputs, including vision, instruction, robot state, and audio context, and outputs a semantic latent $z_{t_k}$ and a key--value cache $\mathrm{KV}_{t_k}$. It also predicts a text stage description $\hat{y}^{\text{json}}_{t_k}$.
    Its low-level model reuses $\mathrm{KV}_{t_k}$ together with the current state and extracts a control feature $u_{t_k}$.
    The \textbf{Advancer} is a decoder-only transformer that predicts a near-future audio code sequence $\mathbf{z}^a_{t_k\rightarrow t_{k+1}}$ from $z_{t_k}$ during training.
    The \textbf{Realizer} synthesizes smooth action chunks via Conditional Flow Matching conditioned on $u_{t_k}$.
    }
    \label{fig:hear_overview}
\end{figure*}

\section{The HEAR Framework}
\label{sec:method}

While the VSLA paradigm establishes the theoretical constraints for sound-centric control, realizing it on physical hardware requires a specific architecture. We present HEAR, an end-to-end framework that instantiates the VSLA paradigm under delayed, low-rate, and asynchronous decision loops. As formalized in Section~\ref{sec:formulation}, sound-centric manipulation highlights two timing-related difficulties. First, short-lived acoustic cues can occur within the effective perception-action gap; depending on window length and alignment, they may either leave the fixed audio window before the next decision or remain only as a brief event embedded in a longer mixed segment (equation~\eqref{eq:evidence_vanishing}). This issue is especially pronounced under open-loop action chunking, where the Blind Execution Interval suppresses perception updates during execution. Second, many sound-centric tasks involve extended monitoring phases with quasi-static vision. In such phases, snapshot observations can be nearly indistinguishable across decisions (equation~\eqref{eq:temporal_aliasing}), and chunked control can become unstable over long horizons consistent with Temporal Motion Collapse~\citep{act}.

HEAR follows a simple systems view to resolve these bottlenecks. Since large backbones cannot be queried at sensor rates in real robots, we decouple high-frequency auditory sensing from low-frequency reasoning and action generation. A lightweight audio buffer is maintained to continuously record transient acoustic events between policy queries. The main policy then operates at a slower control rate, conditioning its decisions on the contents of this buffer. In addition, we introduce a predictive learning signal on near-future audio dynamics, which encourages the shared representation to encode the flow of time even when vision changes little.

HEAR comprises four modules in the following order. The \textbf{Historizer} maintains a causal audio memory that bridges execution gaps (Section~\ref{sec:historizer}). The \textbf{Envisioner} fuses multi-modal inputs into compact representations for control (Section~\ref{sec:envisioner}). The \textbf{Advancer} predicts near-future audio codes from the Envisioner latent to provide temporal grounding during learning (Section~\ref{sec:advancer}). The \textbf{Realizer} then generates smooth action chunks for continuous control (Section~\ref{sec:realizer}). Figure~\ref{fig:hear_overview} illustrates the overall data flow.

\subsection{Architectural Overview}
\label{sec:arch_overview}

We follow the VSLA observation interface in equation~\eqref{eq:obs}. At decision time $t_k$, the freshest sensor data corresponds to the delayed index $\bar{t}_k$ (equation~\eqref{eq:delayed_time}), and the per-decision observation is $o_{t_k} = (I_{\bar{t}_k}^{1:V}, A_{t_k}, l, s_{\bar{t}_k})$. The audio window $A_{t_k}$ contains only the most recent $T_{\text{win}}$ seconds of waveform available by $\bar{t}_k$ (equation~\eqref{eq:audio_window}). When a cue is brief and the next decision is delayed, the cue can fall outside this window under short-window or adverse-alignment settings; under longer windows, it can remain present but occupy only a small fraction of the query-time waveform (equation~\eqref{eq:evidence_vanishing}).

HEAR augments this interface with a persistent audio memory $h_{t_k}$ computed from streaming audio arriving between delayed boundaries $\bar{t}_{k-1}$ and $\bar{t}_k$. The Historizer updates this memory independently of the policy query cadence. The Envisioner then maps the observation together with $h_{t_k}$ into a semantic latent $z_{t_k}$ and a low-dimensional control embedding $u_{t_k}$. During learning, the Advancer predicts near-future audio codes from $z_{t_k}$, which encourages the shared representation to reflect temporal progression under quasi-static vision. Finally, the Realizer generates an $H$-step action chunk from $u_{t_k}$ that is executed open-loop.

We summarize the computation at decision time $t_k$ as:
\begin{equation}
\label{eq:hear_loop}
\begin{aligned}
h_{t_k} &= f_{\varphi}\!\left(h_{t_{k-1}},\, \mathcal{A}_{n_{\bar{t}_{k-1}}+1 \,:\, n_{\bar{t}_k}}\right),\\
(z_{t_k}, u_{t_k}) &= g_{\psi}\!\left(o_{t_k},\, h_{t_k}\right),\\
p_{\eta}\!\left(\mathbf{z}^{a}_{t_k\rightarrow t_{k+1}} \mid z_{t_k}\right) &= \text{Adv}_{\eta}(z_{t_k}),\\
\hat{\mathbf{a}}_k &\sim \pi_{\theta}\!\left(\,\cdot \mid u_{t_k}\right).
\end{aligned}
\end{equation}
Here, $f_{\varphi}$ denotes the streaming audio update, $g_{\psi}$ denotes the Envisioner encoder, $\text{Adv}_{\eta}$ denotes the Advancer network parameterized by $\eta$, and $\pi_{\theta}$ denotes the action-chunk generator.
The predicted chunk $\hat{\mathbf{a}}_k=(\hat{a}_{t_k},\dots,\hat{a}_{t_k+H-1})$ is executed open-loop for $H_{\text{exec}}$ control cycles (Section~\ref{sec:async_inference}).
The Advancer branch provides a predictive learning signal that shapes the latent representation $z_{t_k}$. At deployment, HEAR does not need to explicitly generate the future audio codes $\mathbf{z}^{a}_{t_k\rightarrow t_{k+1}}$, but the temporal structure induced by this objective remains in the latent and benefits control.

\begin{figure}[t]
    \centering
    \includegraphics[width=\linewidth]{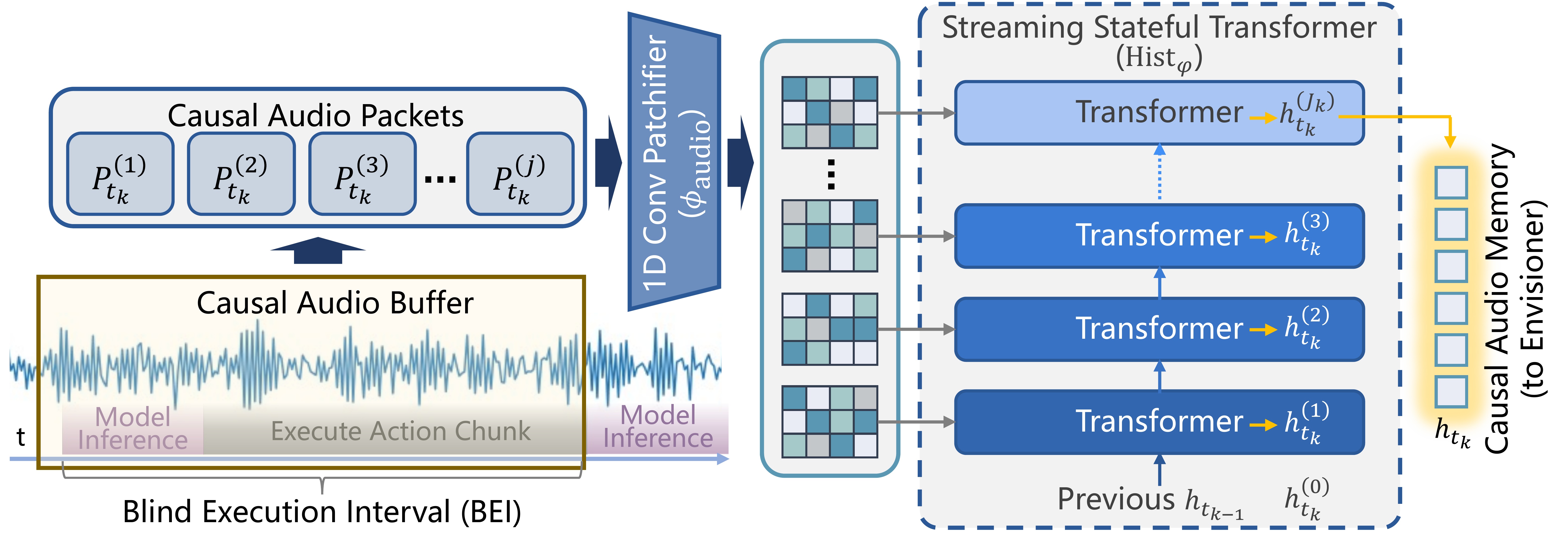}
    \caption{\textbf{The Historizer module.} It processes continuous audio streams into discrete packets and uses a streaming stateful transformer to maintain a persistent causal memory. This mechanism ensures that transient acoustic events occurring during blind execution intervals are preserved for the next decision cycle.}
    \label{fig:Historizer}
\end{figure}

\subsection{Historizer}
\label{sec:historizer}
As shown in Figure~\ref{fig:Historizer}, the Historizer provides causal persistence for transient acoustic cues. In sound-centric manipulation, a brief event can occur while the robot is executing an open-loop action chunk. A fixed query-time audio window may miss the event if the window is short, and even when the window is long enough to contain it, the event can be diluted among unrelated pre- and post-event audio. To keep such evidence available with its causal status intact, the Historizer maintains a compact memory state that is updated from streaming audio independently of the policy query cadence.

\subsubsection{Streaming Stateful Summarization}
We distinguish the fixed per-decision window $A_{t_k}$ (equation~\eqref{eq:audio_window}) from the streaming packets used by the Historizer. At decision time $t_k$, due to end-to-end latency, the freshest audio available to the system corresponds to the delayed boundary $\bar{t}_k$ (equation~\eqref{eq:delayed_time}). We therefore update the memory using the newly arrived samples between successive delayed boundaries, $\mathcal{A}_{n_{\bar{t}_{k-1}}+1 \,:\, n_{\bar{t}_k}}$. We partition this stream into $J_k$ short causal packets $\{P_{t_k}^{(j)}\}_{j=1}^{J_k}$ ordered from old to new, where each packet contains $L$ samples.
To avoid cutting short acoustic events at packet boundaries, adjacent streaming packets are overlapped. The overlap is implemented with a carry-over buffer, so boundary-spanning cues remain visible to the causal audio frontend without changing the delayed-decision interface.
In practice, these packets can be accumulated in a small ring buffer and processed online as they arrive, and the current memory is read out at decision times.

We implement the Historizer as a Streaming Stateful Transformer that carries a compact memory across packets. A lightweight causal audio frontend $\phi_{\text{audio}}$ maps each packet to token features, using a small causal 1D convolutional patchifier.
 Let $q_j$ denote the first sample index of the $j$-th emitted packet in the streaming scheduler. The packet is
\begin{equation}
P_{t_k}^{(j)}=\mathcal{A}_{q_j:q_j+L-1}, \qquad q_{j+1}-q_j=L_s,\quad L_s<L,
\end{equation}
with packets emitted only after the endpoint $q_j+L-1$ has arrived by $\bar{t}_k$. This preserves causality and avoids dropping boundary samples. The readout for decision $t_k$ is the latest memory state after all packets available before $\bar{t}_k$ have been processed, $h_{t_k}=h_{t_k}^{(J_k)}$.
The memory update is
\begin{equation} \label{eq:hist_update}
\begin{aligned}
h^{(0)}_{t_k} &= h_{t_{k-1}},\\
h^{(j)}_{t_k} &= \text{Hist}_{\varphi}\!\left(h^{(j-1)}_{t_k},\, \phi_{\text{audio}}(P_{t_k}^{(j)})\right), \quad j=1,\dots,J_k,\\
h_{t_k} &= h^{(J_k)}_{t_k},
\end{aligned}
\end{equation}
where $\text{Hist}_{\varphi}$ denotes the core transformer block parameterized by $\varphi$, and $h_{t_k}$ is the final causal audio memory provided to the Envisioner at decision time $t_k$. At the beginning of an episode ($k=0$), the initial memory $h_{t_{-1}}$ is set to a zero vector. This streaming process instantiates the general update function $f_{\varphi}$ defined in equation~\eqref{eq:hear_loop}. This update is causal because it depends only on audio that has arrived by $\bar{t}_k$.

\subsubsection{Bridging Execution Gaps}
Because $\{P_{t_k}^{(j)}\}$ covers the audio that arrives between delayed decision boundaries, short-lived cues that occur during open-loop execution can still be incorporated into $h_{t_k}$. This reduces the chance of evidence loss or dilution (equation~\eqref{eq:evidence_vanishing}) by ensuring that the next decision can condition on a persistent summary of what happened acoustically since the previous query.

A practical requirement is that the Historizer has sufficient effective coverage horizon $T_{\text{mem}}$, so that salient evidence remains representable in $h_{t_k}$ until the next decision that needs it. A useful design target is
\begin{equation} \label{eq:tmem_condition}
T_{\text{mem}} \ \ge\ \max_k\!\left(\frac{\Delta_k + \tau_{\text{sys}}}{f_c}\right),
\end{equation}
where $T_{\text{mem}}$ is measured in seconds, $\Delta_k$ and $\tau_{\text{sys}}$ are measured in control cycles, and $f_c$ is the controller frequency in cycles per second. The division by $f_c$ therefore converts the gap into physical time. Thus a transient cue occurring within the effective perception-action gap can still influence the next decision.

Finally, maintaining a memory state allows us to keep the fixed input window $A_{t_k}$ relatively modest. In our experience, relying on a very long window alone can mix distinct interaction phases and make performance more sensitive to the exact window length. By combining a short causal window with a persistent summary $h_{t_k}$, HEAR can retain relevant history while staying aligned with streaming, delayed deployment.

\subsection{Envisioner}
\label{sec:envisioner}
\begin{figure}[t]
    \centering
    \includegraphics[width=\linewidth]{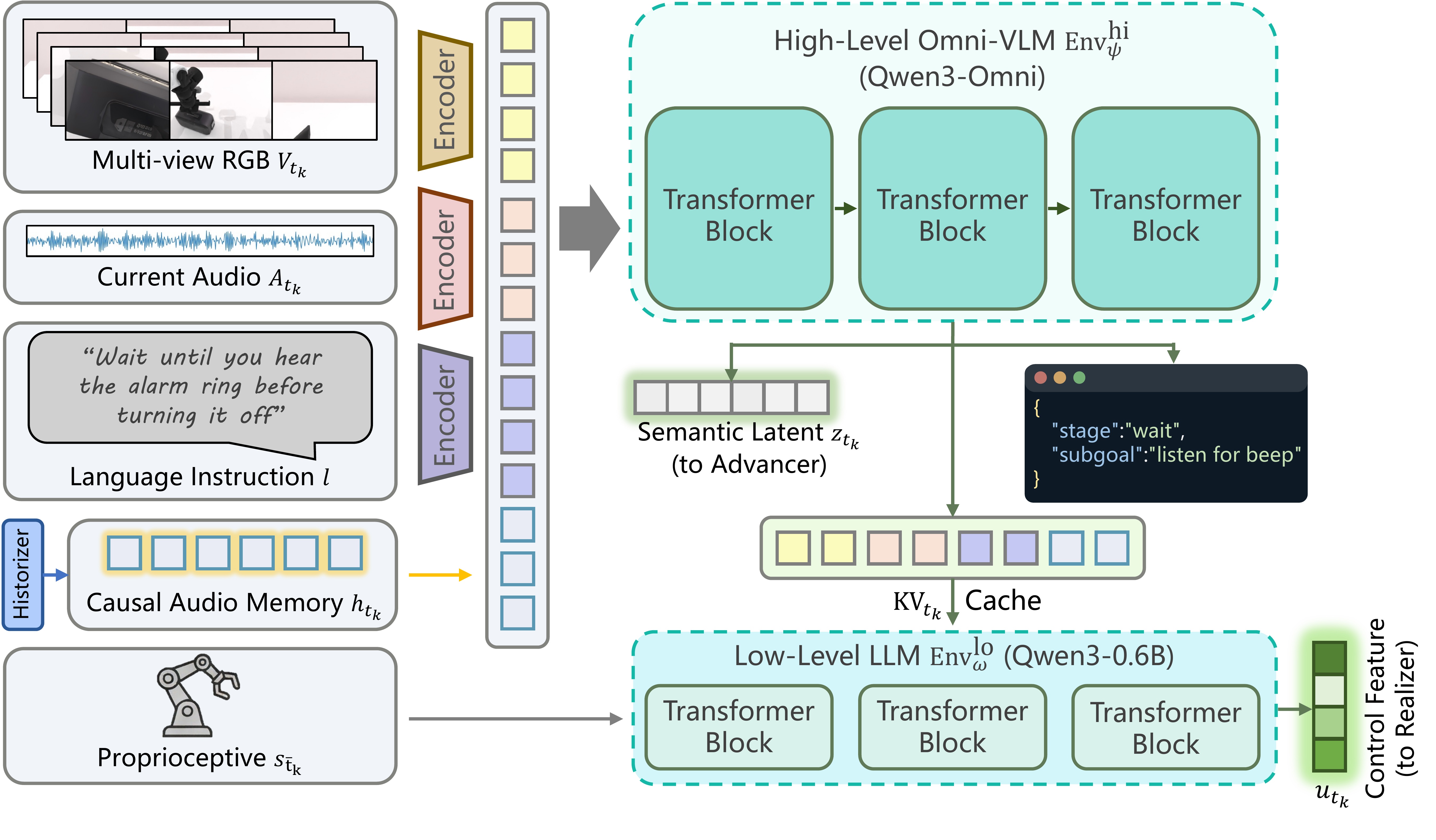}
    \caption{\textbf{The Envisioner module.} A hierarchical reasoning architecture fuses multimodal inputs to guide manipulation. The high-level omni-modal model extracts semantic latents and stage descriptions, while the low-level model reuses the resulting key-value cache alongside proprioceptive data to efficiently generate control features.}
    \label{fig:Envisioner}
\end{figure}

As depicted in Figure~\ref{fig:Envisioner}, the Envisioner serves as the perception-and-reasoning core of HEAR. It adopts a coarse-to-fine hierarchical architecture, implementing the encoder $g_{\psi}$ in equation~\eqref{eq:hear_loop}. At each decision time $t_k$, it fuses the VSLA observation $o_{t_k}$ with the Historizer memory $h_{t_k}$. Its design separates slow multi-modal reasoning from fast control decoding. Concretely, a high-level omni-modal module of Qwen3-Omni~\citep{qwen3omni} summarizes the multi-modal context and produces three outputs: (i) a semantic latent $z_{t_k}$ for temporal learning in the Advancer, (ii) a key--value cache $\mathrm{KV}_{t_k}$ for efficient reuse by a lightweight low-level module of Qwen3-0.6B~\citep{yang2025qwen3}, and (iii) a structured stage description $\hat{y}^{\text{json}}_{t_k}$ generated by the high-level model's text head. A low-level module then conditions on $\mathrm{KV}_{t_k}$ together with the current robot state to produce a low-dimensional control feature $u_{t_k}$ for the Realizer.

\subsubsection{High-Level Omni-LLM}
\label{sec:envisioner_hi}
We denote the high-level module as $\text{Env}^{\text{hi}}_{\psi}$. Its inputs follow the VSLA interface:
\begin{itemize}
    \item Multi-view RGB images $V_{t_k}\triangleq I_{\bar{t}_k}^{1:V}$.
    \item Natural language instruction $l$.
    \item Robot proprioceptive state $s_{\bar{t}_k}$.
    \item The current audio window $A_{t_k}$ (equation~\eqref{eq:audio_window}).
    \item The causal audio memory $h_{t_k}$ from the Historizer.
\end{itemize}

The module produces a semantic latent $z_{t_k}$ and a KV cache $\mathrm{KV}_{t_k}$ for reuse:
\begin{equation}
\label{eq:env_hi}
(z_{t_k}, \mathrm{KV}_{t_k}) = \text{Env}^{\text{hi}}_{\psi}\!\left(V_{t_k}, A_{t_k}, l, s_{\bar{t}_k}, h_{t_k}\right).
\end{equation}
In addition, leveraging its native text generation capability, it outputs a structured stage record $\hat{y}^{\text{json}}_{t_k}$ at the same time.

\subsubsection{Stage Decomposition}
\label{sec:stage_decomp}
Sound-centric tasks often contain visually similar phases with different intent, such as holding still to monitor versus acting after a trigger. To make this progress explicit, the high-level module generates a short structured stage description $\hat{y}^{\text{json}}_{t_k}$, for example \texttt{\{"stage":"wait","subgoal":"listen for beep"\}}. We train this output by minimizing token-level negative log-likelihood over the JSON string. We denote this auxiliary objective as $\mathcal{L}_{\text{text}}$. This loss encourages the semantic latent $z_{t_k}$ to track progress and to separate phases that may be visually quasi-static but acoustically distinct.

\subsubsection{Low-Level LLM}
\label{sec:envisioner_lo}
The low-level module $\text{Env}^{\text{lo}}_{\omega}$ produces a control feature vector $u_{t_k}$ by reusing the cached context $\mathrm{KV}_{t_k}$ together with the robot state:
\begin{equation}
\label{eq:low_level}
u_{t_k} = \text{Env}^{\text{lo}}_{\omega}\!\left(\mathrm{KV}_{t_k}, s_{\bar{t}_k}\right).
\end{equation}
This design reuses the high-level multi-modal context while keeping the control pathway lightweight. The resulting $u_{t_k}$ serves as the conditioning signal for the Realizer, which generates the action chunk executed after time $t_k$.

\subsection{Advancer}
\label{sec:advancer}
\begin{figure}[t]
    \centering
    \includegraphics[width=\linewidth]{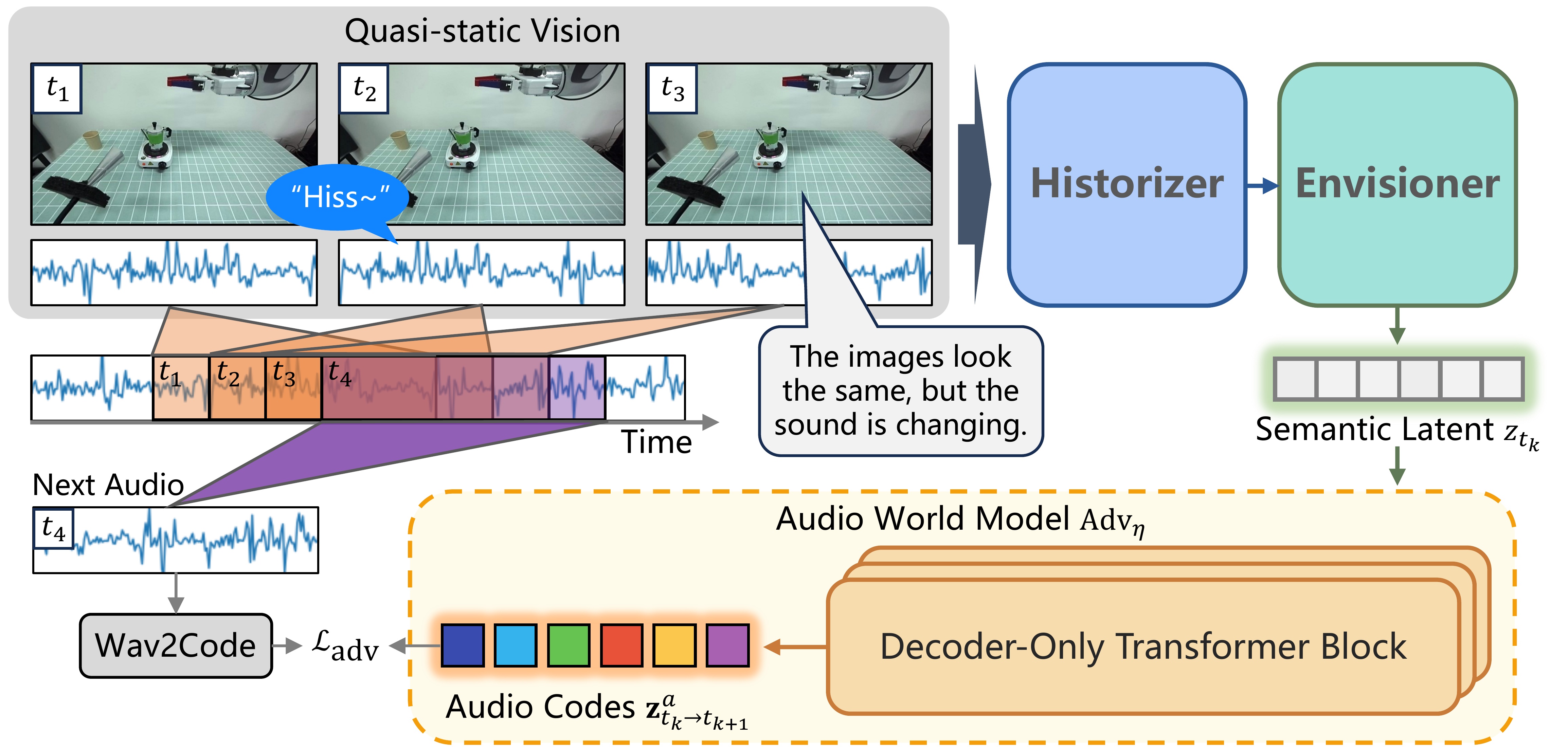}
    \caption{\textbf{The Advancer module.} In scenarios requiring sustained waiting with quasi-static visual observations, this decoder-only audio world model predicts near-future acoustic codes. This predictive objective grounds the shared latent representation in continuous time, helping the policy maintain stability and temporal awareness.}
    \label{fig:Advancer}
\end{figure}

Capturing a transient cue is necessary, but long-horizon monitoring also requires a stable notion of time. When vision is quasi-static, a policy that only sees snapshot observations may receive nearly identical inputs across many decisions, even though the desired behavior changes with elapsed time (equation~\eqref{eq:temporal_aliasing}). The Advancer addresses this by learning audio dynamics through near-future prediction, as shown in Figure~\ref{fig:Advancer}. This predictive learning encourages the shared latent $z_{t_k}$ to reflect temporal progression, which can improve stability during extended listening phases and sharpen reactivity once a cue occurs.

\subsubsection{Near-Future Audio Codes}
Rather than predicting raw waveforms, we predict discrete audio codes obtained from a neural audio tokenizer. Let $\text{Tok}(\cdot)$ map a waveform segment to a sequence of discrete code indices. We use a causal streaming tokenizer so that the codes for a segment depend only on samples within that segment, matching the constraints of streaming deployment~\citep{defossez2024moshi}.

We define the prediction target as the code sequence for the audio that arrives after the delayed boundary $\bar{t}_k$ and up to the next delayed boundary $\bar{t}_{k+1}$:
\begin{equation}
\label{eq:audio_token_target}
\mathbf{z}^{a}_{t_k\rightarrow t_{k+1}}
\triangleq
\text{Tok}\!\left(\mathcal{A}_{n_{\bar{t}_k}+1 \,:\, n_{\bar{t}_{k+1}}}\right).
\end{equation}
This definition makes explicit that the target corresponds to a time interval, rather than a single instant. Note that while we denote the target with the decision interval subscript $t_k \rightarrow t_{k+1}$ for simplicity, the actual prediction interval is strictly aligned with the delayed boundaries $\bar{t}_k$ and $\bar{t}_{k+1}$ to respect causality. Because $\text{Tok}$ is causal, the codes for this interval depend only on samples within the interval, which is consistent with streaming constraints.

\subsubsection{Latent Prediction}
The Advancer, denoted as $\text{Adv}_{\eta}$, is implemented as a decoder-only predictor parameterized by $\eta$ that takes $z_{t_k}$ as input and outputs a distribution over the future code sequence:
\begin{equation}
\label{eq:adv_pred}
p_{\eta}\!\left(\mathbf{z}^{a}_{t_k\rightarrow t_{k+1}} \mid z_{t_k}\right)
= \text{Adv}_{\eta}(z_{t_k}).
\end{equation}
We train it with cross-entropy over codebook indices:
\begin{equation}
\label{eq:adv_loss}
\mathcal{L}_{\text{adv}} =
- \mathbb{E}_{\mathcal{D}}
\left[\log p_{\eta}\!\left(\mathbf{z}^{a}_{t_k\rightarrow t_{k+1}}\mid z_{t_k}\right)\right].
\end{equation}
where $\mathcal{D}$ denotes the training dataset. At deployment, HEAR does not need to explicitly generate $\mathbf{z}^{a}_{t_k\rightarrow t_{k+1}}$. The role of the Advancer is to inject a concrete temporal learning signal into the shared latent, so that $z_{t_k}$ better tracks time-evolving process states when vision changes little.

\subsection{Realizer}
\label{sec:realizer}
The Realizer maps the low-level control feature $u_{t_k}$ to an $H$-step joint-position action chunk $\hat{\mathbf{a}}_k=(\hat{a}_{t_k},\dots,\hat{a}_{t_k+H-1})$. The robot then executes this chunk open-loop for $H_{\text{exec}}$ control cycles, as described in Section~\ref{sec:async_inference}. We model the Realizer as a conditional generator,
\begin{equation}
\hat{\mathbf{a}}_k \sim \pi_{\theta}\!\left(\,\cdot \mid u_{t_k}\right).
\end{equation}

In sound-centric manipulation, smooth chunks are useful for tracking accuracy, and they can also make listening more reliable. Abrupt motion may introduce mechanical ego-noise and transient vibration that mask weak acoustic cues, especially during monitoring phases~\citep{maniwav}.

We instantiate $\pi_{\theta}$ using conditional flow matching (CFM)~\citep{flow_matching, liu2022flow}. CFM learns a continuous vector field that transports samples from a simple Gaussian prior to the action-chunk distribution along a straight-line path. This structure is convenient for real robots because it can generate chunks with a small, fixed number of integration steps. Compared with diffusion-style action generators~\citep{diffusion_policy}, this can reduce the inference overhead in low-rate decision loops.

\subsubsection{Conditioning}
We condition the flow network on $u_{t_k}$ using adaptive normalization. Concretely, a small MLP maps $u_{t_k}$ to per-block scale and shift parameters, and these parameters modulate the normalized activations~\citep{perez2018film}. This provides a lightweight way to adapt the generated action distribution to the inferred task context~\citep{flow_matching}.

\subsubsection{Vector Field Regression}
Let $\mathbf{a}_k^*$ be the expert action chunk of length $H$ at decision time $t_k$ (equation~\eqref{eq:bc_target}). We flatten the chunk into a vector $x_1 \triangleq \mathrm{vec}(\mathbf{a}_k^*) \in \mathbb{R}^{H d_{\text{qpos}}}$, and sample Gaussian noise $x_0 \sim \mathcal{N}(0,I)$. For flow time $\lambda\in[0,1]$, we define the linear transport path $x_\lambda = (1-\lambda)x_0 + \lambda x_1$.

The Realizer learns a conditional vector field $v_{\xi}(x_\lambda,\lambda,u_{t_k})$ and regresses it to the target direction $(x_1-x_0)$:
\begin{equation}
\label{eq:flow_loss}
\mathcal{L}_{\text{flow}} \!=\!
\mathbb{E}_{\substack{\mathcal{D},\, x_0\sim\mathcal{N}(0,I) \\ \lambda\sim\mathcal{U}(0,1)}}
\!\!\left[\left\|v_{\xi}(x_\lambda, \lambda, u_{t_k}){-}(x_1{-}x_0)\right\|_2^2\right]\!,
\end{equation}
where $u_{t_k}$ is produced by the low-level Envisioner from $(\mathrm{KV}_{t_k}, s_{\bar{t}_k})$ (equation~\eqref{eq:low_level}).

\subsubsection{Inference via Euler Integration}
At test time, we sample $x_0$ and numerically integrate the ODE defined by the learned vector field:
\begin{equation}
\frac{d x_\lambda}{d\lambda} = v_{\xi}(x_\lambda, \lambda, u_{t_k}),
\end{equation}
from $\lambda=0$ to $\lambda=1$. To satisfy real-time constraints, we use Euler integration with a fixed number of steps. The solution at $\lambda=1$ yields $\hat{x}_1$, which we reshape into the predicted action chunk $\hat{\mathbf{a}}_k$ and execute open-loop for $H_{\text{exec}}$ cycles.

\subsection{Training Objectives}
\label{sec:training}
We train HEAR with imitation learning from expert action chunks. Each training sample contains the per-decision observation $o_{t_k}$ (equation~\eqref{eq:obs}), the expert action chunk $\mathbf{a}_k^*$ (equation~\eqref{eq:bc_target}), and a stage label $y_{t_k}^{\text{json}*}$. We denote the dataset of extracted training samples as $\mathcal{D} = \{(o_{t_k}, \mathcal{A}_{n_{\bar{t}_{k-1}}+1 \,:\, n_{\bar{t}_{k+1}}}, \mathbf{a}_k^*, y_{t_k}^{\text{json}*})\}_{k=1}^{N}$, which explicitly includes the continuous audio segments required for the Historizer memory update and the Advancer target.

The total objective is a weighted sum:
\begin{equation}
\mathcal{L}_{\text{total}}
=
\mathcal{L}_{\text{flow}}
+
w_{\text{adv}}\,\mathcal{L}_{\text{adv}}
+
w_{\text{text}}\,\mathcal{L}_{\text{text}},
\end{equation}
where $\mathcal{L}_{\text{flow}}$ trains the Realizer for action-chunk generation (equation~\eqref{eq:flow_loss}), $\mathcal{L}_{\text{adv}}$ trains the Advancer for near-future audio code prediction (equation~\eqref{eq:adv_loss}), and $\mathcal{L}_{\text{text}}$ supervises the structured stage description generated by the high-level Envisioner (Section~\ref{sec:stage_decomp}).

In practice, we gradually increase $w_{\text{adv}}$ and $w_{\text{text}}$ during early training. This lets the model first learn basic action imitation, and then place more emphasis on temporal
grounding and progress tracking.

\section{Data and Benchmark Suite}
\label{sec:data_benchmark}
VSLA research requires both audio-aligned robot data and evaluation protocols that enforce sound causality. Existing large-scale VLA datasets provide visual observations, language, proprioception, and actions, but rarely include synchronized microphone streams. Standard manipulation benchmarks likewise evaluate final geometric success rather than whether the action was causally triggered by sound. We therefore introduce two resources: OpenX-Sound, an audio-augmented Open X-Embodiment subset for scalable robotics-specific pretraining, and HEAR-Bench, a sound-centric simulation benchmark that evaluates delayed, chunked policies under strict timing constraints.

\subsection{OpenX-Sound: Audio-Augmented Open X-Embodiment}
\label{sec:openx_sound}
The original Open X-Embodiment (OpenX) dataset~\citep{openx} provides broad task and embodiment diversity but no microphone recordings. Collecting synchronized robot audio at comparable scale would require substantial hardware standardization across laboratories. OpenX-Sound is therefore constructed as a practical pretraining resource: selected OpenX trajectories are augmented with time-aligned audio tracks so that the VSLA policy stack can learn the robotic coupling among sound, vision, proprioception, and actions before task-specific fine-tuning. Because the base Qwen3-Omni backbone already has broad multimodal pretraining on general audio-video data, OpenX-Sound is used as a robotics-specific temporal-alignment stage rather than as generic multimodal pretraining. It is not intended to replace real robot audio.

For every selected OpenX episode, we preserve the original multi-view RGB, language instructions, proprioception, and expert actions, while adding the synthesized audio track. The detailed OpenX-Sound scale and format statistics are provided in Supplemental Section~S2 together with episode selection, video-to-audio synthesis, alignment review, metadata packaging, and domain-gap limitations.

Because downstream sound-causal tasks are timing-sensitive, OpenX-Sound passes a review workflow before release. In this workflow, all episodes undergo human inspection, supported by automated alignment and metadata checks. Episodes with semantically inconsistent audio, implausible event timing, or obvious synchronization errors are corrected or removed. The dataset is released through the project page and Hugging Face.

\subsection{HEAR-Bench: Sound-Centric Simulation Benchmark}
\label{sec:hear_bench}
While OpenX-Sound supports pretraining, evaluation requires tasks in which the correctness of an action depends on acoustic timing. Standard manipulation benchmarks (e.g., RLBench~\citep{rlbench}, ManiSkill~\citep{maniskill2}) primarily test spatial reasoning and visual control in tasks without task-relevant audio. HEAR-Bench is a sound-enabled manipulation benchmark built on RoboTwin2~\citep{robotwin2} (Figure~\ref{fig:robot_setups}(a)). It provides a streaming microphone channel and enforces the sound-causal success rule in equation~\eqref{eq:timed_success}.

\textbf{Simulation platform and audio pipeline.}
HEAR-Bench is an event-conditioned audio simulator: it models when task-relevant sounds enter the stream and what their dominant acoustic content is. A dedicated asynchronous audio thread mixes task-defined event clips into a background stream, and each policy query receives only the causal audio window $A_{t_k}$. The full playback pipeline, sample-index alignment, dual-arm control interface, and anti-shortcut randomization protocol are described in Supplemental Section~S3.

\paragraph{A Taxonomy of Manipulation Sounds.}
To systematically evaluate auditory awareness, we organize the decision-critical acoustic cues in HEAR-Bench into four primary categories derived from everyday human experience: human speech, event-triggered alarms, continuous process sounds, and physical interaction feedback. We instantiate these categories across seven distinct tasks, as detailed in Table~\ref{tab:hearbench_tasks}.

\begin{table}[ht]
\small\sf\centering
\caption{\textbf{HEAR-Bench tasks.} The benchmark evaluates four primary categories of decision-critical acoustic cues under strict temporal constraints.}
\label{tab:hearbench_tasks}
\begin{tabular}{ll}
\toprule
Task & Primary cue \\
\midrule
\multicolumn{2}{l}{\textit{\textbf{Event-triggered alarms}}} \\
\task{Alarm Clock} & alarm ringing \\
\task{Microwave} & completion beep \\
\addlinespace
\multicolumn{2}{l}{\textit{\textbf{Human speech}}} \\
\task{Check Yes} & prosody-driven speech \\
\task{Interrupt} & spoken interruption \\
\addlinespace
\multicolumn{2}{l}{\textit{\textbf{Continuous process sounds}}} \\
\task{Pour Water} & pouring sound evolution \\
\task{Boil Water} & boiling sound evolution \\
\addlinespace
\multicolumn{2}{l}{\textit{\textbf{Physical interaction feedback}}} \\
\task{Check Materials} & contact acoustics \\
\bottomrule
\end{tabular}
\end{table}

Across all tasks, success is strictly governed by equation~\eqref{eq:timed_success}. If the robot reaches the geometric goal before the relevant acoustic cue has occurred, the trial is recorded as a failure. We randomize initial visual states and cue timings to prevent policies from memorizing visual sequences or exploiting spurious correlations. By combining short-lived cues and long waiting phases, HEAR-Bench directly tests whether a policy can bridge sensory execution gaps in real time. Additional task-construction and audio-playback details are provided in Supplemental Section~S3.

\section{Experiments}
\label{sec:experiments}
We evaluate HEAR in simulation and on physical hardware. The simulation setting uses HEAR-Bench on the dual-arm RoboTwin2 platform (Figure~\ref{fig:robot_setups}a). The real-world setting deploys the policy on a single-arm Franka Panda (Figure~\ref{fig:robot_setups}b). In both settings, the primary metric is the sound-causal success rate defined in Section~\ref{sec:timed_success}: a trial is successful only when the robot completes the physical goal after the relevant acoustic condition has occurred.

The predicate $g(\mathcal{H}^{\text{audio}}_t)$ is instantiated according to the natural acoustic precondition of each task. For onset tasks, such as \task{Alarm Clock}, \task{Microwave}, and \task{Real Alarm Clock}, the cue is the first reliable alarm, beep, or ring onset. For speech tasks, such as \task{Check Yes}, \task{Interrupt}, and the dialogue phases of \task{Answer Phone}, the cue is the relevant utterance or dialogue boundary. For continuous processes, such as \task{Pour Water}, \task{Boil Water}, and \task{Moka Coffee}, the cue is the task-defined sustained acoustic state. For active-probing tasks, such as \task{Check Materials} and \task{Shake Bottle}, the cue is the robot-generated contact or shaking sound. These instantiations differ by task, but they enforce the same causal rule in equation~\eqref{eq:timed_success}: completion before the relevant acoustic evidence is invalid.

Our experimental design addresses three questions. First, can a policy use diverse acoustic cues, including transient impacts, process sounds, and prosody, for closed-loop manipulation? Second, can it preserve cues that occur during delayed or low-rate decision loops? Third, which architectural components are responsible for performance under these timing constraints? The tasks isolate acoustic evidence through randomized cue timing, visually ambiguous states, and long monitoring phases in which the relevant progress variable is carried primarily by sound.

\subsection{Experimental Setup}
\label{sec:exp_setup}

\subsubsection{Observation Interface and Action Space}
\label{sec:exp_interface}
All evaluated methods operate strictly under the formal VSLA observation interface defined in equation~\eqref{eq:obs}. At decision time $t_k$, the policy receives multi-view RGB images $I_{\bar{t}_k}^{1:V}$, a causal audio window $A_{t_k}$, a natural language instruction $l$, and proprioception $s_{\bar{t}_k}$. The audio input is consistently provided as a 1D raw waveform vector sampled from a single microphone, ensuring the external interface remains uniform even if specific baselines compute spectrograms internally. Proprioception consists of the robot's joint-position state (qpos), encompassing all controllable joints and gripper degrees of freedom. At each inference step $t_k$, the policy generates an action chunk of joint-position targets $\hat{\mathbf{a}}_{k} \in \mathbb{R}^{H \times d_{\text{qpos}}}$. To ensure a controlled comparison, we maintain a consistent chunk horizon of $H=16$ across all methods unless explicitly stated otherwise.

Following the minimalist design of recent end-to-end systems \citep{pi0}, our control pipeline is highly simplified. The predicted action chunks encode target joint and gripper positions, which are unrolled and executed open-loop by a simple low-level controller, without relying on any additional trajectory planning, inverse kinematics solvers, or collision detection.

\begin{figure}[t]
\centering
\includegraphics[width=\linewidth]{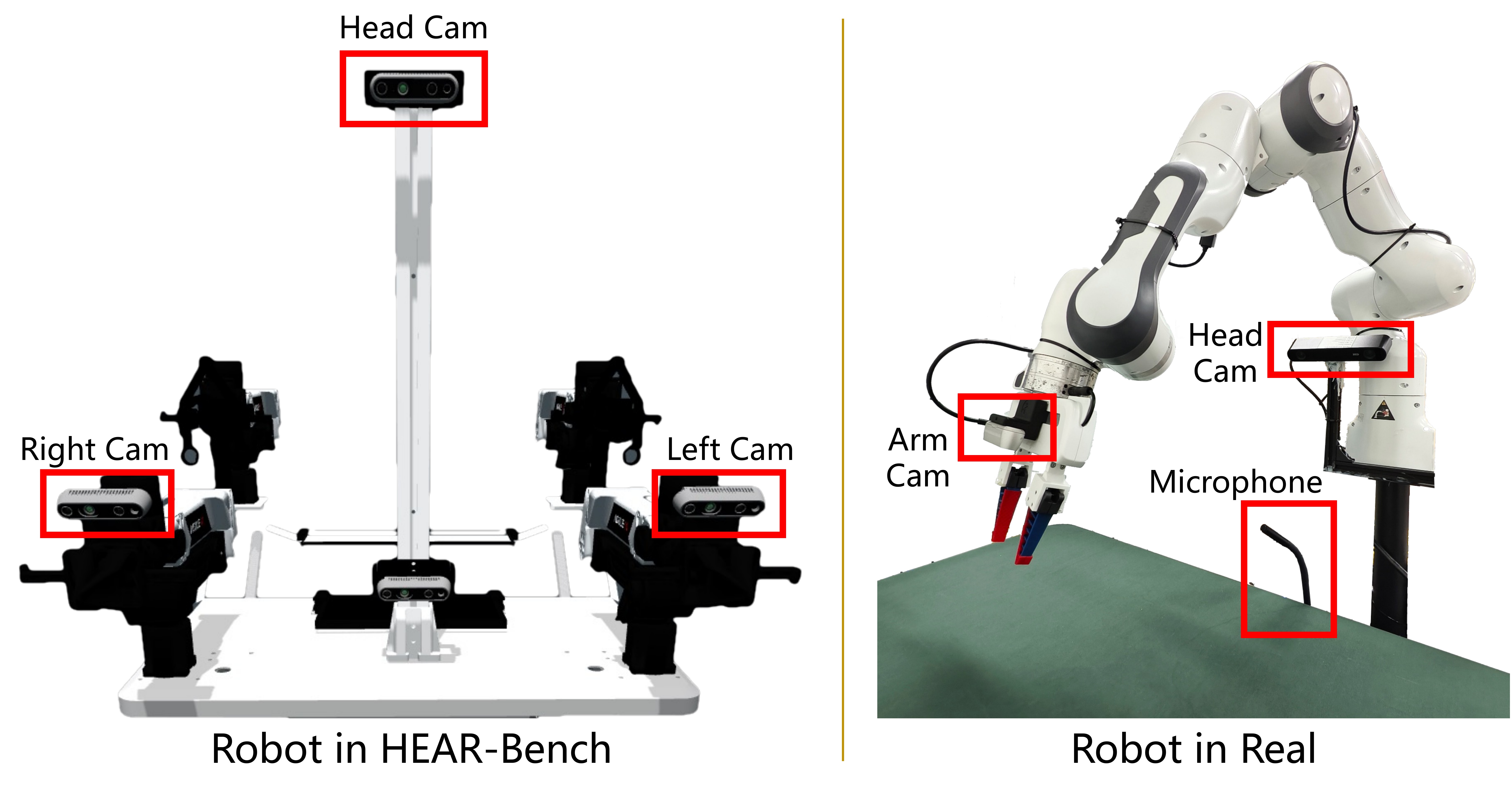}
    \caption{\textbf{Robot platforms and sensor suites.} \textbf{(a)} For HEAR-Bench simulations, we use a dual-arm RoboTwin2 equipped with three RGB camera views (Head Cam, Left Cam, Right Cam). \textbf{(b)} For real-world experiments, we deploy a single-arm Franka Panda with a head-mounted camera, an arm-mounted camera, and a standard workspace microphone.}

    \label{fig:robot_setups}
\end{figure}

We evaluate two distinct robot configurations (Figure~\ref{fig:robot_setups}). In the HEAR-Bench simulation, we utilize the dual-arm RoboTwin2 platform equipped with three RGB camera views (head, left, and right). For real-world deployments, we use a single-arm Franka Panda equipped with a head-mounted RGB camera, an arm-mounted RGB camera, and a standard workspace microphone.

\subsubsection{Training Protocol and Implementation Details}
\label{sec:exp_protocol_impl}
All policies are trained via offline imitation learning using a two-stage recipe: an initial pretraining phase on the OpenX-Sound dataset (Section~\ref{sec:openx_sound}) to establish foundational multi-sensory representations, followed by task-specific fine-tuning on either HEAR-Bench or real-robot demonstrations. We evaluate each task over 100 independent trials, randomizing initial visual scenes and cue timings to prevent policies from memorizing fixed temporal schedules.
The training data are collected using the ground-truth target states and cue schedules. In both HEAR-Bench and the Franka setup, rule-based controllers and motion planners produce sound-causal trajectories, while synchronized RGB, proprioception, actions, and microphone streams are recorded for imitation learning.

For evaluation, the task-specific predicate $g$ in equation~\eqref{eq:timed_success} is instantiated by cue type rather than by a universal amplitude threshold. Single-onset tasks become eligible after the relevant alarm, beep, ring, or contact onset; speech and dialogue tasks use the corresponding utterance or end-of-dialogue boundary; continuous-process tasks such as \task{Pour Water}, \task{Boil Water}, and \task{Moka Coffee} become eligible when the task-defined sustained acoustic state is reached; and active-probing tasks such as \task{Check Materials} and \task{Shake Bottle} require the robot-generated interaction sound. Thus a trial can only be counted as successful after the appropriate acoustic condition for that task has occurred.

The evaluated HEAR implementation follows the module definitions in Section~\ref{sec:method}: an SST-based Historizer, a hierarchical Envisioner, a training-time Advancer, and a flow-matching Realizer for qpos action chunks. The detailed architectural sizes, audio preprocessing settings, optimizer schedule, training horizons, loss weights, hardware configuration, and deployed timing logs are provided in Supplemental Sections~S4 and S5.11. For real-robot experiments, the end-to-end latency $\tau_{\text{sys}}$ is empirically estimated from timestamped sensing and command logs, whereas in simulation we default to $\tau_{\text{sys}}=0$ to explicitly isolate the execution gaps induced by action chunking. On the Franka deployment, the measured executed action-chunk interval is approximately 2.1\,s and the effective BEI is approximately 2.3\,s, so policies need to act at the next decision boundary after a cue has been preserved.

\subsubsection{Baselines}
\label{sec:exp_baselines}
We compare HEAR with VLA backbones, audio adapters, and compact audio-native policies. All methods use the same two-stage training protocol, receive the same visual observations and causal audio segments when applicable, and output the same qpos action chunks.

\paragraph{VLA Backbones and Audio Adapters.}
We use \textbf{OpenVLA}~\citep{openvla} and \textbf{$\pi_{0.5}$}~\citep{pi0} as representative VLA backbones. Their default variants receive only $I_{\bar{t}_k}^{1:V}$, $l$, and $s_{\bar{t}_k}$. We add two common audio adapters. The \textbf{-Waveform} variant renders the causal waveform window $A_{t_k}$ as a 2D line plot and passes it as an additional camera view. This preserves coarse amplitude structure but collapses temporal and spectral detail into a static image. The \textbf{-ASR} variant transcribes the audio and appends the transcript to the language prompt, which is useful for explicit speech but discards non-speech cues and prosody.

\paragraph{Compact Audio-Native Policies.}
We compare with \textbf{Play it by Ear}~\citep{du2022playitbyear} and \textbf{ManiWAV}~\citep{maniwav}, which natively ingest audio for audio-visual manipulation. These policies are useful baselines because native audio input and faster inference can reduce some timing pressure. Their compact architectures, however, do not explicitly preserve low-level acoustic evidence across delayed chunked execution while supporting long-horizon multimodal reasoning.

\subsection{Simulation on HEAR-Bench}
\label{sec:exp_sim}
We rigorously evaluate our approach in simulation using HEAR-Bench (Section~\ref{sec:hear_bench}), built upon the dual-arm RoboTwin2 platform (Figure~\ref{fig:robot_setups}a). HEAR-Bench provides a streaming microphone signal and strictly enforces the sound-causal success rules defined in equation~\eqref{eq:timed_success}. The fundamental difficulty of this benchmark lies in bridging execution gaps without missing brief acoustic events, maintaining stable control during prolonged, visually quasi-static waiting phases, and understanding physical contact sounds.

\subsubsection{Task Suite and Design Rationale}
\label{sec:exp_sim_tasks}
We systematically design seven tasks in HEAR-Bench to cover a complementary taxonomy of sound-centric challenges (Table~\ref{tab:hearbench_tasks}). Figures~\ref{fig:hearbench_alarmclock_scene} through \ref{fig:hearbench_boil_water_scene} provide representative qualitative key frames from HEAR rollouts in simulation.
\begin{itemize}
    \item \textbf{Sustained trigger, \task{Alarm Clock}.} Tests indefinite waiting and suppressing visually plausible early actions until an auditory precondition is satisfied.
    \item \textbf{Impulse trigger, \task{Microwave}.} Tests capturing a brief event that can occur inside an execution gap (especially the Blind Execution Interval under open-loop chunking), then acting after the cue has ceased and may be diluted inside a later query-time window.
    \item \textbf{Prosody, \task{Check Yes}.} Uses identical text but different intonation to force reasoning beyond ASR outputs.
    \item \textbf{Contact and materials, \task{Check Materials}.} Uses visually similar objects so the impact timbre serves as a low-cost proxy for physical property sensing.
    \item \textbf{Continuous monitoring, \task{Pour Water} and \task{Boil Water}.} Requires tracking gradual acoustic evolution over long horizons when vision is weak or static.
    \item \textbf{Random override, \task{Interrupt}.} Places a spoken interrupt at random times to discourage ignoring audio and to test safe mode switching mid-motion.
\end{itemize}

\begin{table*}
\scriptsize\sf\centering
\caption{\textbf{Simulation results on HEAR-Bench.} Success rates $\uparrow$ are averaged over 100 independent trials per task. The ``VLA'' column indicates whether the baseline relies on a vision-language-action backbone.}
\setlength{\tabcolsep}{0.9mm}
\label{tab:sim_results}
\begin{tabular}{l|c|cccccccc}
\toprule
Method & VLA & \task{Alarm Clock} & \task{Check Yes} & \task{Check Materials} & \task{Pour Water} & \task{Boil Water} & \task{Microwave} & \task{Interrupt} & Avg.\\
\midrule
OpenVLA~\citep{openvla} & Y & 0.00 & 0.33 & 0.31 & 0.00 & 0.00 & 0.00 & 0.03 & 0.10\\
OpenVLA-Waveform~\citep{openvla} & Y & 0.71 & 0.60 & 0.63 & 0.26 & 0.37 & 0.45 & 0.69 & 0.53\\
OpenVLA-ASR~\citep{openvla} & Y & 0.16 & 0.59 & 0.10 & 0.06 & 0.03 & 0.41 & 0.65 & 0.29\\
$\pi_{0.5}$~\citep{pi0} & Y & 0.00 & 0.48 & 0.43 & 0.00 & 0.00 & 0.00 & 0.09 & 0.14\\
$\pi_{0.5}$-Waveform~\citep{pi0} & Y & 0.80 & 0.73 & 0.71 & 0.31 & 0.43 & 0.52 & 0.75 & 0.61\\
$\pi_{0.5}$-ASR~\citep{pi0} & Y & 0.21 & 0.70 & 0.15 & 0.09 & 0.04 & 0.52 & 0.74 & 0.35\\
Play it by Ear~\citep{du2022playitbyear} & N & 0.45 & 0.38 & 0.39 & 0.15 & 0.10 & 0.12 & 0.38 & 0.28\\
ManiWAV~\citep{maniwav} & N & 0.40 & 0.31 & 0.41 & 0.12 & 0.17 & 0.08 & 0.32 & 0.26\\
\textbf{HEAR} & Y & \textbf{0.91} & \textbf{0.89} & \textbf{0.83} & \textbf{0.51} & \textbf{0.81} & \textbf{0.85} & \textbf{0.88} & \textbf{0.81}\\
\bottomrule
\end{tabular}
\end{table*}

\subsubsection{Main Results}
\label{sec:exp_sim_main}
Table~\ref{tab:sim_results} reports the success rates over 100 trials per task. HEAR achieves the highest overall performance, demonstrating an average success rate of 81\%. Among the VLA baselines, rendering a waveform as an image provides a moderate improvement for tasks with distinct volume spikes, whereas the ASR adapter primarily benefits tasks featuring explicit speech commands.

Notably, the compact audio-native policies, Play it by Ear and ManiWAV, outperform the vision-only backbones but generally trail behind the adapted VLA models and HEAR. Because these lightweight models run at higher inference rates, the gap is not primarily a runtime issue; their compact architectures have less capacity for long-horizon multimodal state tracking. Vision-only policies perform poorly on this sound-gated suite, as expected, because they cannot observe the causal trigger. Under an identical training protocol, OpenVLA and its variants consistently underperform $\pi_{0.5}$ and its variants, indicating that sound-centric closed-loop manipulation remains sensitive to backbone capacity and action modeling choices.

\subsubsection{Trigger Tasks}
\label{sec:exp_sim_triggers}

\begin{figure}[h]
\centering
\includegraphics[width=\linewidth]{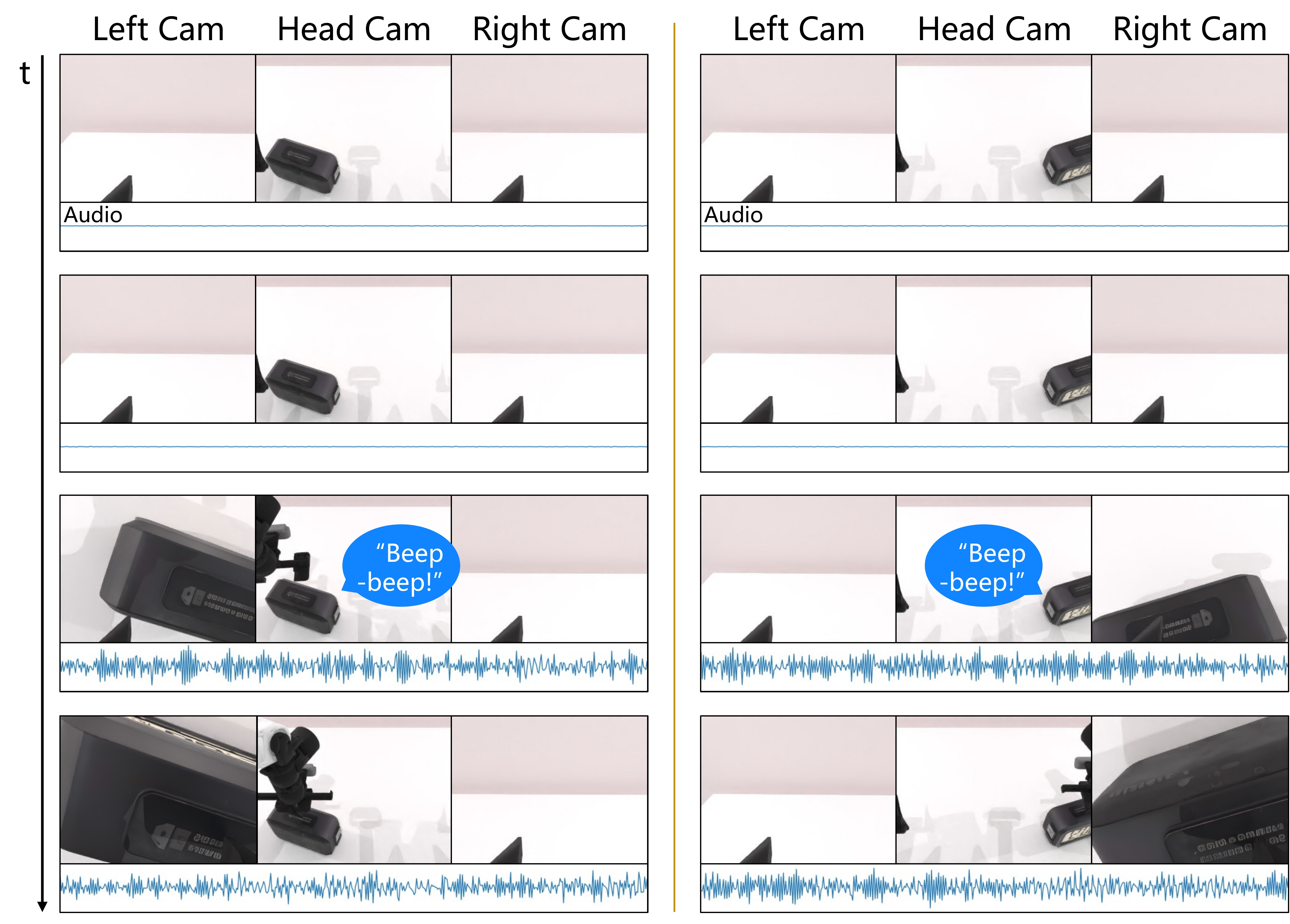}
\caption{\textbf{Qualitative execution on HEAR-Bench: \task{Alarm Clock}.} The policy waits for the ring onset, autonomously selects an arm, and presses the stop button strictly after the acoustic cue. This prevents visually plausible but premature actions.}
\label{fig:hearbench_alarmclock_scene}
\end{figure}

\paragraph{\task{Alarm Clock}.}
For the \task{Alarm Clock} task, the robot is positioned in front of a clock whose stop button remains visible throughout the episode, as shown in Figure~\ref{fig:hearbench_alarmclock_scene}. The alarm begins ringing at a random time. The robot must keep both arms still until the ringing starts and press the button only after the onset, leaving arm selection to the policy. Pressing before the ring violates sound-causality and is counted as a failure under equation~\eqref{eq:timed_success}.

This setup is intentionally designed as a tempting shortcut. A vision-only policy can always see the button and may attempt to press it immediately, which would appear reasonable under standard goal-reaching metrics. Randomizing the onset time prevents fixed-delay heuristics and highlights the asynchronous nature of acoustic evidence. Because the onset can fall between inference cycles and be weakly represented by a one-shot windowed audio interface, success requires robustness to temporal misalignment.

As shown in Table~\ref{tab:sim_results}, HEAR achieves a 0.91 success rate. The best baseline is $\pi_{0.5}$-Waveform at 0.80, followed by OpenVLA-Waveform at 0.71. ASR-based variants perform poorly since the cue is non-speech. Vision-only policies perform poorly as expected, confirming that visual heuristics alone are insufficient when critical state transitions are acoustically gated. When baselines fail, the dominant error is an early press driven by visual affordances. Waveform rendering helps once the ring is fully present because the amplitude spike is visually obvious, but it often relies on threshold-like patterns and remains sensitive to where the onset falls inside the rendered window. HEAR avoids these artifacts by encoding the raw 1D waveform and integrating onset evidence into a causal memory, allowing the policy to wait stably and commit only after the cue.

\begin{figure}[h]
\centering
\includegraphics[width=\linewidth]{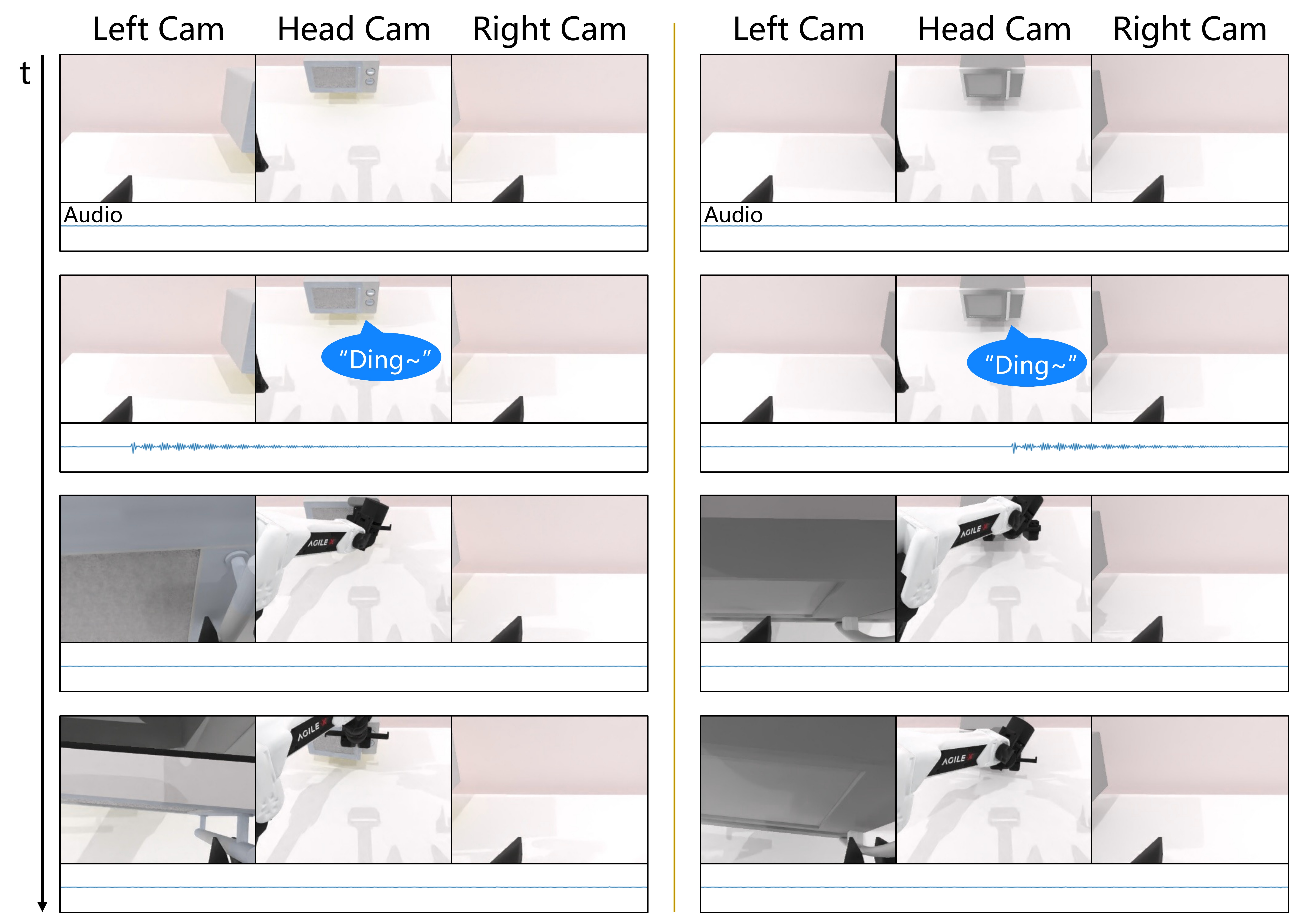}
\caption{\textbf{Qualitative execution on HEAR-Bench: \task{Microwave}.} A brief ``ding'' occurs at a randomized time. The robot successfully captures this transient cue and opens the door only after the sound has finished.}
\label{fig:hearbench_microwave_scene}
\end{figure}

\paragraph{\task{Microwave}.}
The \task{Microwave} scenario requires the robot to stand by until the appliance emits a brief ``ding'' at a randomized time, as shown in Figure~\ref{fig:hearbench_microwave_scene}. The correct behavior is to open the door only after this transient cue has occurred.

This task is designed to stress transient-cue preservation under execution gaps. The acoustic cue can be shorter than the effective perception-action gap $G_k$ (equation~\eqref{eq:decision_latency}) and may occur while the policy is not being queried, such as inside the Blind Execution Interval under open-loop chunking (Section~\ref{sec:async_inference}). Under short-window or unfavorable-alignment settings it can be absent from the next query window; under the deployed window it is often present only as a low-duty-cycle event inside a longer segment, which makes timing and causal relevance difficult for memoryless policies (equation~\eqref{eq:evidence_vanishing}).

HEAR maintains a 0.85 success rate. The strongest baselines are $\pi_{0.5}$-Waveform and $\pi_{0.5}$-ASR, both reaching 0.52. Audio-native baselines are also limited on this task, with Play it by Ear achieving 0.12 and ManiWAV achieving 0.08. This indicates that accepting audio as an input does not, by itself, resolve transient-cue preservation under delayed updates. Waveform image rendering fails primarily when the short amplitude spike is poorly aligned with the rendered window, causing the robot to wait indefinitely. HEAR performs well because the Historizer module compresses transient evidence into a persistent causal memory state $h_{t_k}$ (equation~\eqref{eq:vsla_policy_history}), enabling correct actions even after the sound has ceased.

\subsubsection{Spoken Commands and Prosody}
\label{sec:exp_sim_speech}

\begin{figure}[h]
\centering
\includegraphics[width=\linewidth]{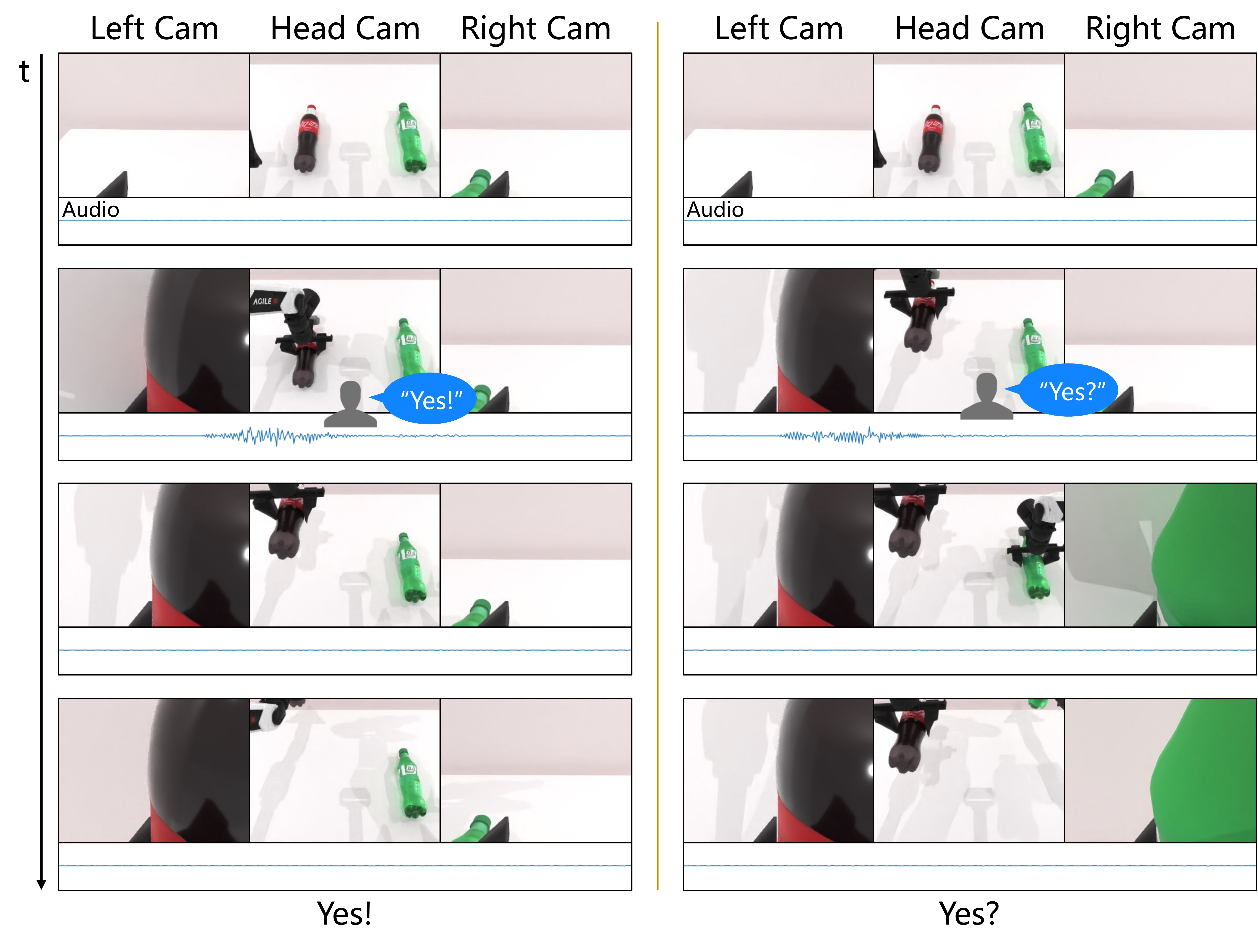}
\caption{\textbf{Qualitative execution on HEAR-Bench: \task{Check Yes}.} The spoken transcript is consistently ``Yes''. The robot relies entirely on intonation and prosody to determine whether to hold the current object or switch to the alternative.}

\label{fig:hearbench_check_yes_scene}
\end{figure}

\paragraph{\task{Check Yes}.}
To evaluate prosody comprehension, \task{Check Yes} presents the robot with two identical bottles and requires it to identify the correct one, as shown in Figure~\ref{fig:hearbench_check_yes_scene}. The episode is constructed so the robot first lifts the initial bottle and then receives spoken feedback that is lexically identical but prosodically different. ``Yes!'' indicates affirmation, meaning the robot picked the correct bottle. ``Yes?'' indicates doubt, meaning it must switch to the other bottle.

This task requires speech understanding beyond basic transcripts. A pure ASR-to-text interface collapses both outcomes into near-identical token sequences, discarding the pragmatic intent. Robust success therefore benefits from modeling intonation directly from the waveform, a capability critical for real human-robot interaction where prosody can invert meaning.

HEAR reaches a 0.89 success rate. While ASR captures the lexical content, it compresses intonation into text, blurring the distinction between affirmation and doubt ($\pi_{0.5}$-ASR achieves 0.70). The waveform-as-image fails because pitch and prosody variations are largely invisible in a standard amplitude envelope plot. HEAR improves upon these baselines by learning a dedicated 1D waveform encoder that preserves prosody and fusing these features directly with vision, language, and proprioception.

\begin{figure}[h]
\centering
\includegraphics[width=\linewidth]{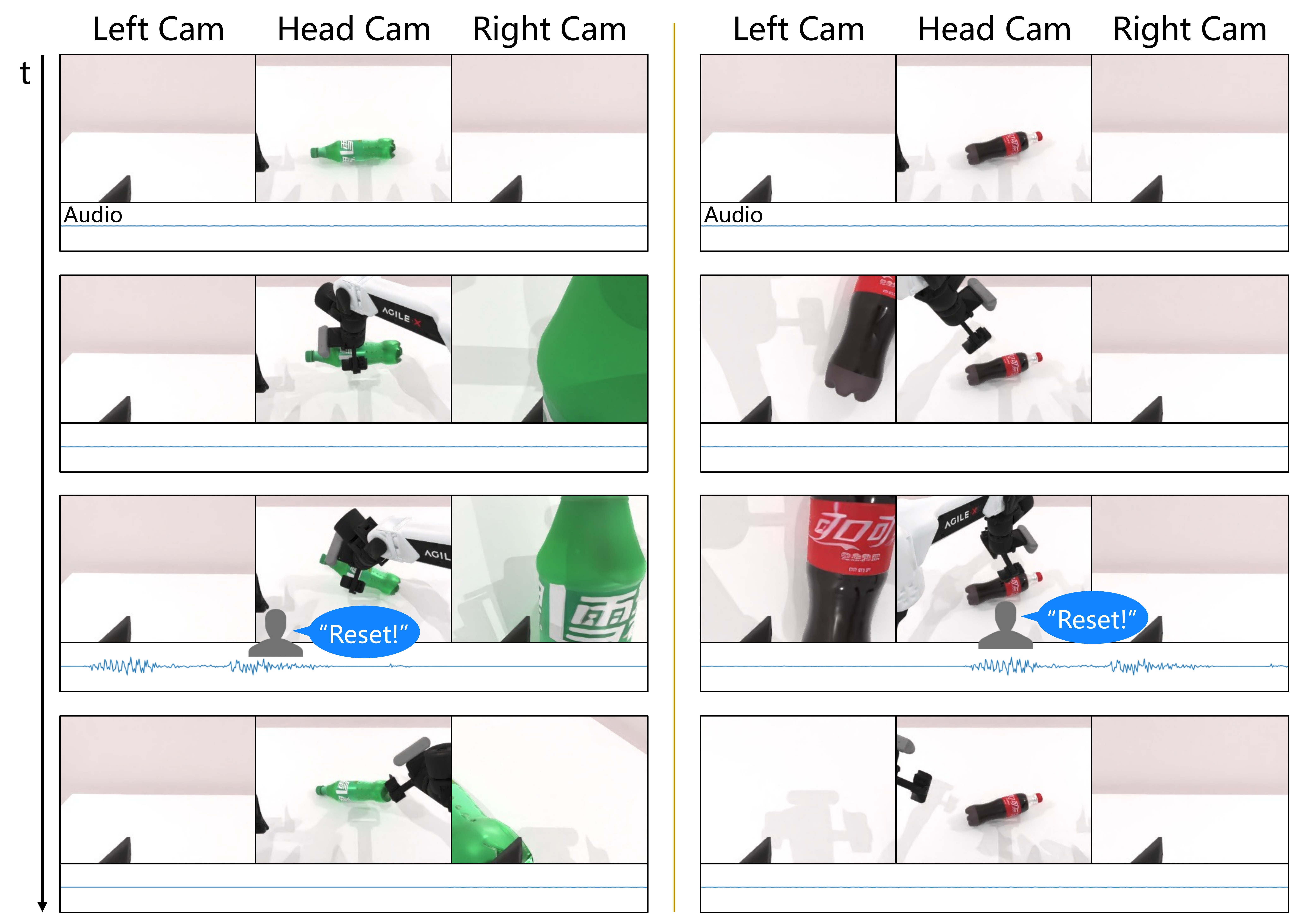}
\caption{\textbf{Qualitative execution on HEAR-Bench: \task{Interrupt}.} While performing an ongoing manipulation routine, the robot receives a randomized spoken ``reset'' command. It overrides its current motion and safely returns to the initial pose.}
\label{fig:hearbench_interrupt_scene}
\end{figure}

\paragraph{\task{Interrupt}.}
Beyond understanding intent, robots must also respond to speech dynamically during execution. During the \task{Interrupt} evaluation, as illustrated in Figure~\ref{fig:hearbench_interrupt_scene}, the robot initiates a standard manipulation routine, autonomously choosing which arm to use (Figure~\ref{fig:hearbench_interrupt_scene}). At a random time mid-motion, a spoken ``reset'' command is issued. The correct response is to terminate the current trajectory and return to the initial pose.

Because the interrupt timing is randomized, an ordinary visual policy that completes the routine without listening will fail under sound-causality metrics. Furthermore, this task stresses action binding, as the same keyword must trigger different low-level control outputs depending on the robot's current joint positions.

HEAR leads with a 0.88 success rate. Waveform and ASR baselines perform reasonably well (0.75 and 0.74 for $\pi_{0.5}$, respectively) because the cue is explicit speech. However, reacting safely under chunked control depends heavily on how the command is fused with the current motion state. HEAR improves robustness by integrating speech evidence into the same representation used for action chunk generation, producing smoother joint-position trajectories that minimize overshoot during abrupt course corrections.

\subsubsection{Contact and Materials}
\label{sec:exp_sim_materials}
\begin{figure}[h]
\centering
\includegraphics[width=\linewidth]{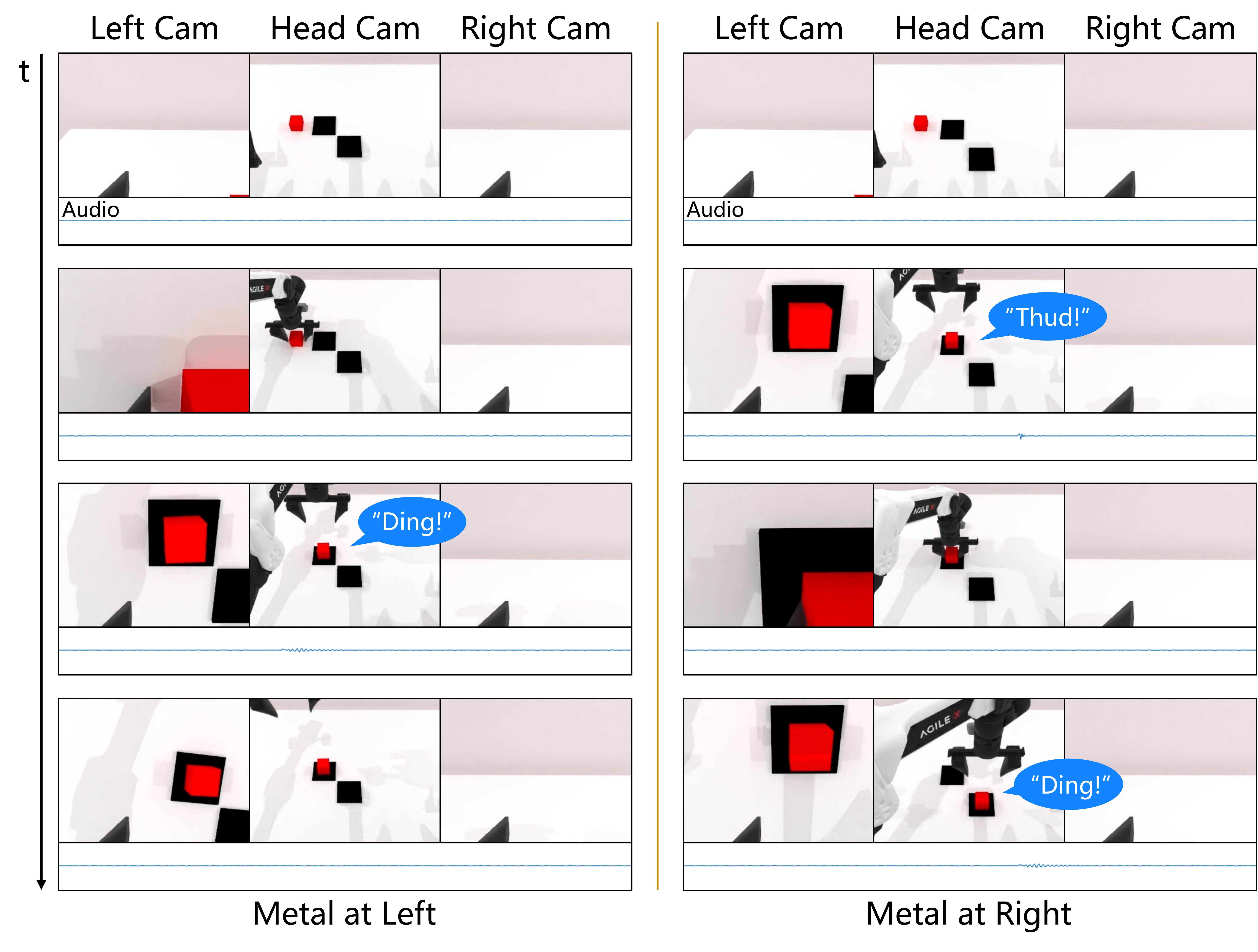}
\caption{\textbf{Qualitative execution on HEAR-Bench: \task{Check Materials}.} The robot drops a cube onto one of two visually identical plates. It infers the underlying material property entirely from the resulting impact acoustics to complete the sorting task.}
\label{fig:hearbench_check_materials_scene}
\end{figure}

\paragraph{\task{Check Materials}.}
As depicted in Figure~\ref{fig:hearbench_check_materials_scene}, the \task{Check Materials} setup features two visually identical plates placed within the workspace. One plate is the target material, and the other is a distractor. The robot must pick up a cube, drop it onto the first plate to produce an impact sound, and infer from the audio whether it landed on the target material. If incorrect, it must move the cube to the second plate.

This task serves as a controlled example of using hearing as a low-cost physical sensor. Material properties often require tactile feedback to disambiguate when vision is aliased~\citep{clarke2018learning}. Here, we make vision intentionally uninformative to force reliance on contact timbre. Unlike \task{Alarm Clock}, the robot must actively generate informative audio and then deliberately hold still to listen, tightly coupling perception quality to control stability.

HEAR achieves a 0.83 success rate. ASR variants are not well matched to this setting (0.15 for $\pi_{0.5}$-ASR), because impact acoustics cannot be converted into useful text. We observe two dominant issues in baseline failures. First, memoryless policies are sensitive to the exact placement of the brief impact within the observation window. Second, waveform images are limited in capturing the subtle frequency differences (timbre) between materials, making the visual plot highly ambiguous. HEAR benefits from operating on the raw audio stream and retaining causal memory for the transient impact and smoother chunk generation through the Realizer module, which helps the robot pause deliberately and increases the effective signal-to-noise ratio.

\subsubsection{Process Monitoring}
\label{sec:exp_sim_process}
\begin{figure}[h]
\centering
\includegraphics[width=\linewidth]{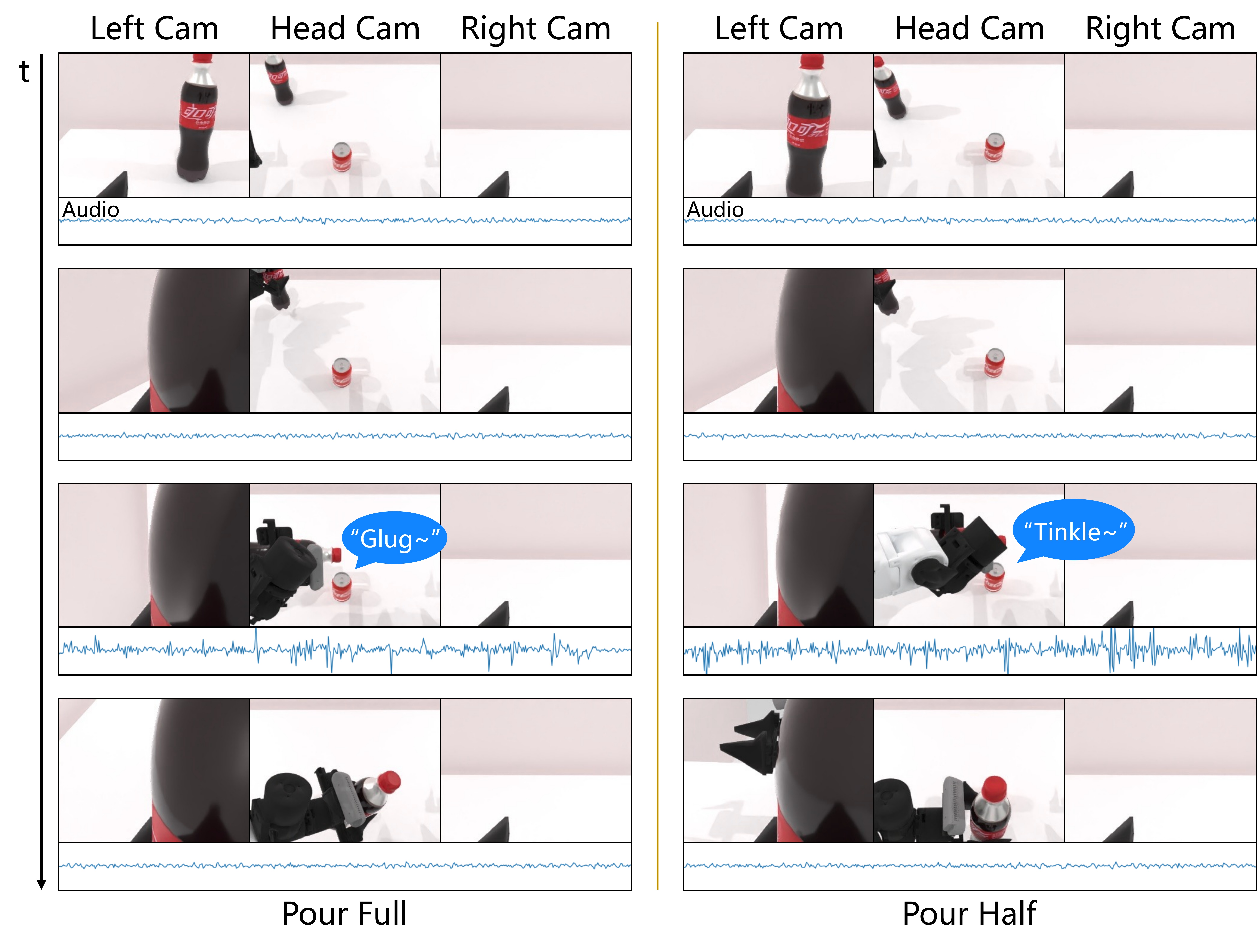}

\caption{\textbf{Qualitative execution on HEAR-Bench: \task{Pour Water}.} The robot dispenses liquid into an opaque container and stops precisely at a target level guided by continuous acoustic feedback. The vertical axis of the audio waveform is scaled by 10x for visualization.}
\label{fig:hearbench_pour_water_scene}
\end{figure}

\paragraph{\task{Pour Water}.}
When executing \task{Pour Water}, the robot dispenses liquid into an opaque container and must halt precisely at a target level specified by the language instruction, a process visualized in Figure~\ref{fig:hearbench_pour_water_scene}. Since the fill level is visually occluded, the decision signal is the continuous evolution of the pouring sound.

We include this task to evaluate continuous temporal flow. The relevant information is not an isolated acoustic event but the rate and direction of change in sound over time. This challenges snapshot-style audio interfaces because adjacent observation windows can sound remarkably similar even as the correct control decision evolves.

HEAR reaches a 0.51 success rate, outperforming $\pi_{0.5}$-Waveform (0.31). A visual plot of pouring noise often appears as a dense, static block of lines, making it difficult for a vision backbone to track gradual acoustic evolution. In HEAR, the Historizer aggregates evidence over time, but continuous processes also require sensitivity to progression. The Advancer module provides predictive dynamics supervision by predicting near-future acoustic tokens (equations~\eqref{eq:adv_pred} and~\eqref{eq:adv_loss}). This mechanism forces the fused representation to encode local temporal gradients, effectively learning how the sound is changing over time rather than just recognizing what the sound is at a single instant.

\begin{figure}[h]
\centering
\includegraphics[width=\linewidth]{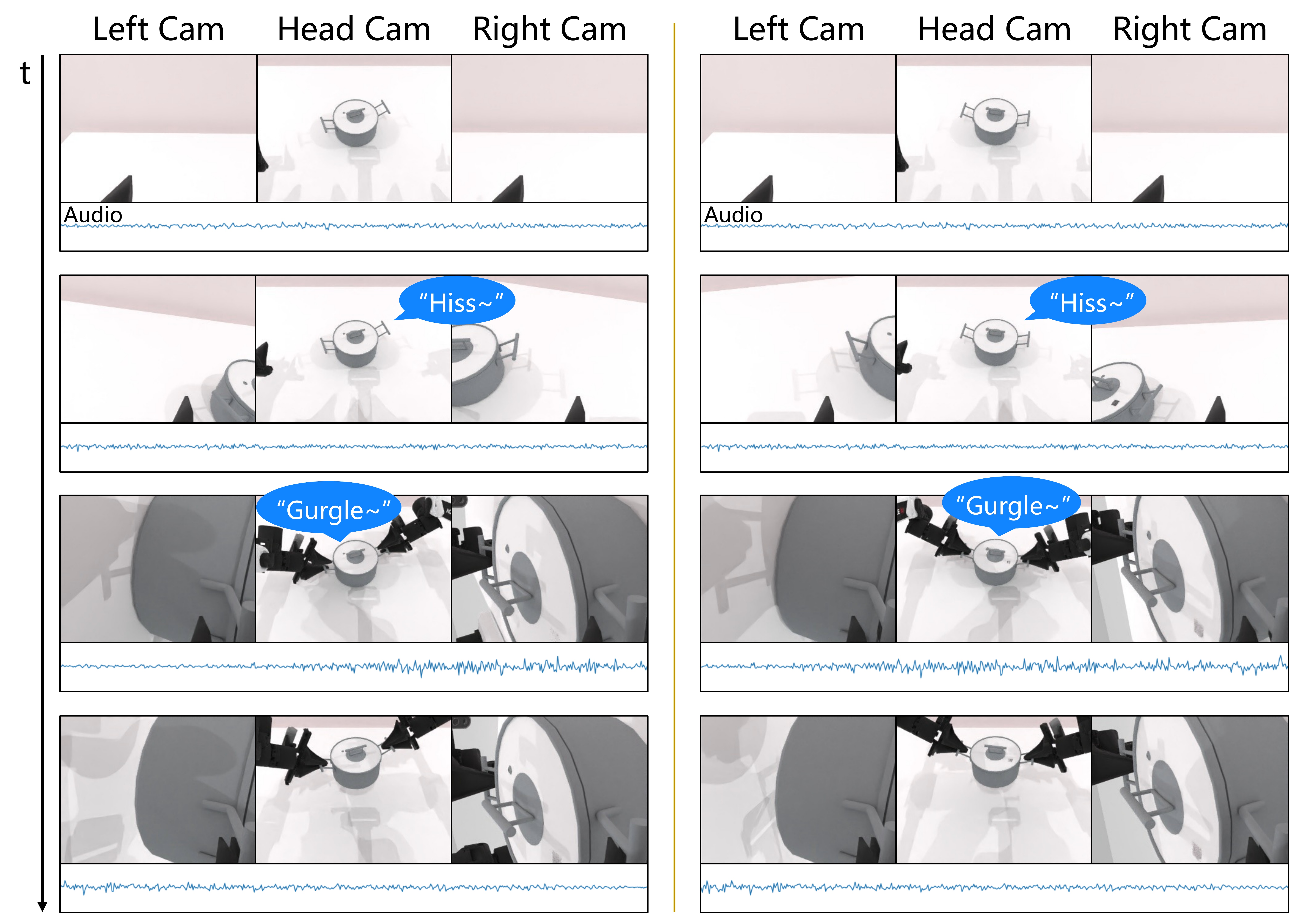}
\caption{\textbf{Qualitative execution on HEAR-Bench:\task{Boil Water}.} The robot maintains a stable posture through a prolonged, visually static period and intervenes strictly once the evolving sound indicates a rolling boil. The vertical axis of the audio waveform is scaled by 10x for visualization.}
\label{fig:hearbench_boil_water_scene}
\end{figure}

\paragraph{\task{Boil Water}.}
Figure~\ref{fig:hearbench_boil_water_scene} demonstrates the \task{Boil Water} environment, which requires the robot to monitor a pot and intervene strictly once the acoustic profile indicates a rolling boil. The time-to-boil is randomized to prevent reliance on fixed waiting schedules.

This task isolates long waiting periods characterized by subtle acoustic transitions. The visual scene remains static while the sound evolves gradually from silence to a simmer, and finally to a sustained boil. Success requires the policy to remain still for an arbitrary duration and detect a sustained acoustic pattern rather than a transient burst.

HEAR achieves a 0.81 success rate, compared to 0.43 for $\pi_{0.5}$-Waveform. Baseline failures typically involve early actions triggered by visually plausible heuristics or overreactions to transient noisy bursts. HEAR mitigates these errors by smoothing short-term fluctuations through causal memory and tracking gradual acoustic progression via the predictive dynamics supervision in Advancer.

\subsubsection{Summary}
\label{sec:exp_sim_summary}
Overall, HEAR demonstrates consistent improvements across diverse sound categories in HEAR-Bench. While waveform rendering is a strong baseline, it remains sensitive to window placement and lacks a principled mechanism for causal history. ASR is effective when cues can be cleanly tokenized into explicit speech, but it discards non-speech details and prosody. HEAR succeeds across both discrete triggers and continuous processes because it treats the raw waveform as a first-class observation modality. The simulation results therefore support the central design choice of combining causal auditory persistence with progression-sensitive temporal modeling. The real-world experiments below test the same mechanism under physical noise, microphone placement, and longer task horizons.

\subsection{Real-Robot Experiments}
\label{sec:exp_real}

\subsubsection{Platform and Acoustic Challenges}
\label{sec:exp_real_platform}
We deploy the trained policies on a single-arm Franka Panda with two RGB views (head and arm cameras) and a single standard microphone placed near the workspace (Figure~\ref{fig:robot_setups}(b)). The policy observes the identical $o_{t_k}$ interface defined in equation~\eqref{eq:obs}. We intentionally avoid specialized acoustic sensors such as contact microphones or ear-in-hand arrays, requiring the system to process raw, ambient acoustic signals.

Real-world audio introduces structural challenges absent in simulation, including room reverberation, non-stationary background noise, and device-specific acoustic variations. Additionally, the robot's own physical motion generates mechanical ego-noise. These factors make stable waiting and deliberate, repeatable contact behaviors far more critical than in pristine simulated environments. All tasks are evaluated over 100 independent trials using the sound-causal success definitions outlined in Section~\ref{sec:timed_success}.

\begin{table*}
\small\sf\centering
\caption{\textbf{Real-robot evaluation results.} Success rates $\uparrow$ across four physical tasks, averaged over 100 trials each. All tasks strictly enforce the sound-causal success criteria, penalizing visually plausible but premature actions. For the two hardest long-horizon tasks, the acoustic setup and task-specific data configuration are synchronized across methods, and all methods were re-evaluated under the same conditions.}
\label{tab:real_results}
\setlength{\tabcolsep}{1.2mm}
\begin{tabular}{l|c|ccccc}
\toprule
Method & VLA & \task{Moka Coffee} & \task{Answer Phone} & \task{Shake Bottle} & \task{Real Alarm Clock} & Avg.\\
\midrule
OpenVLA~\citep{openvla} & Y & 0.00 & 0.00 & 0.22 & 0.01 & 0.06\\
OpenVLA-Waveform~\citep{openvla} & Y & 0.13 & 0.10 & 0.41 & 0.77 & 0.35\\
OpenVLA-ASR~\citep{openvla} & Y & 0.05 & 0.05 & 0.19 & 0.63 & 0.23\\
$\pi_{0.5}$~\citep{pi0} & Y & 0.00 & 0.01 & 0.48 & 0.04 & 0.13\\
$\pi_{0.5}$-Waveform~\citep{pi0} & Y & 0.19 & 0.17 & 0.54 & 0.90 & 0.45\\
$\pi_{0.5}$-ASR~\citep{pi0} & Y & 0.12 & 0.11 & 0.21 & 0.78 & 0.31\\
Play it by Ear~\citep{du2022playitbyear} & N & 0.18 & 0.19 & 0.42 & 0.85 & 0.41\\
ManiWAV~\citep{maniwav} & N & 0.17 & 0.20 & 0.45 & 0.81 & 0.41\\
\textbf{HEAR} & Y & \textbf{0.51} & \textbf{0.43} & \textbf{0.88} & \textbf{0.96} & \textbf{0.70}\\

\bottomrule
\end{tabular}
\end{table*}

\subsubsection{Task Suite and Design Rationale}
\label{sec:exp_real_tasksuite}
We evaluate four real-world tasks that mirror the simulation requirements while explicitly introducing the complexities of physical acoustics. Figures~\ref{fig:real_moka_coffee_scene}--\ref{fig:real_real_alarmclock_scene} provide representative qualitative key frames.
\begin{itemize}
    \item \textbf{\task{Moka Coffee}} (Long-horizon process monitoring). Requires waiting and reacting to subtle acoustic phase changes with minimal visual progression. While previous robotic coffee-making demonstrations typically involve operating automated machines via discrete actions (e.g., pressing a button to dispense)~\citep{fu2024mobile}, this task requires the robot to monitor the continuous, physical stovetop brewing process and make decisions based entirely on auditory feedback.
    \item \textbf{\task{Answer Phone}} (Multi-stage speech and events). Requires responding to a transient ring, understanding speech during a call, and using audio to determine when to end the interaction under severe visual aliasing (repeated visual states).
    \item \textbf{\task{Shake Bottle}} (Self-generated cue). Requires active motion to elicit sound for hidden-state classification, emphasizing the coupling between deliberate proprioception and acoustic perception.
    \item \textbf{\task{Real Alarm Clock}} (Sustained trigger). Tests robust waiting and reaction timing under acoustic domain shifts, including reverberation and background noise.
\end{itemize}

\subsubsection{Results}
\label{sec:exp_real_results}
Table~\ref{tab:real_results} reports the success rates across physical deployments. Under a uniform acoustic setup, HEAR achieved a 0.54 average success rate across the four tasks. We report this deployment-sensitivity reference because it reveals that \task{Moka Coffee} and \task{Answer Phone} are strongly affected by acoustic capture quality and dialogue coverage. For the calibrated comparison, the improved acoustic setup and task-specific data configuration for these two tasks were synchronized across all methods, and every method was evaluated under the same conditions. HEAR reaches a 0.70 average success rate, while the strongest VLA adapter baseline, $\pi_{0.5}$-Waveform, reaches 0.45. Supplemental Table~S7 reports the perturbation-retention analysis using the calibrated clean real-robot setup as the reference.

The audio-native non-VLA policies, Play it by Ear and ManiWAV, remain limited in these complex real-world settings, averaging 0.41. Although these compact policies can theoretically execute at higher frequencies, their smaller parameter scale limits long-horizon acoustic state tracking under real-world variability. Consistent with simulation findings, OpenVLA and its variants underperform $\pi_{0.5}$ and its variants under the same training protocol.

The two hardest real-world tasks, \task{Moka Coffee} and \task{Answer Phone}, expose different practical bottlenecks. For \task{Moka Coffee}, the decisive sputtering signal is weak, so moving the ordinary workspace microphone closer to the pot and applying a simple gain calibration improves HEAR from 0.18 to 0.51 within the same single-microphone audio-input interface. For \task{Answer Phone}, the main difficulty is distinguishing conversational content and the end-of-call moment; expanding the fine-tuning demonstrations with more dialogue variations improves success from 0.15 to 0.43. These task-specific settings are applied consistently to all methods in Table~\ref{tab:real_results}, so the updated comparison reflects both the improved physical setup and the remaining gap between causal auditory memory and native-audio or waveform baselines.

\subsubsection{Task Breakdown}
\label{sec:exp_real_tasks}

\begin{figure}[t]
\centering
\includegraphics[width=\linewidth]{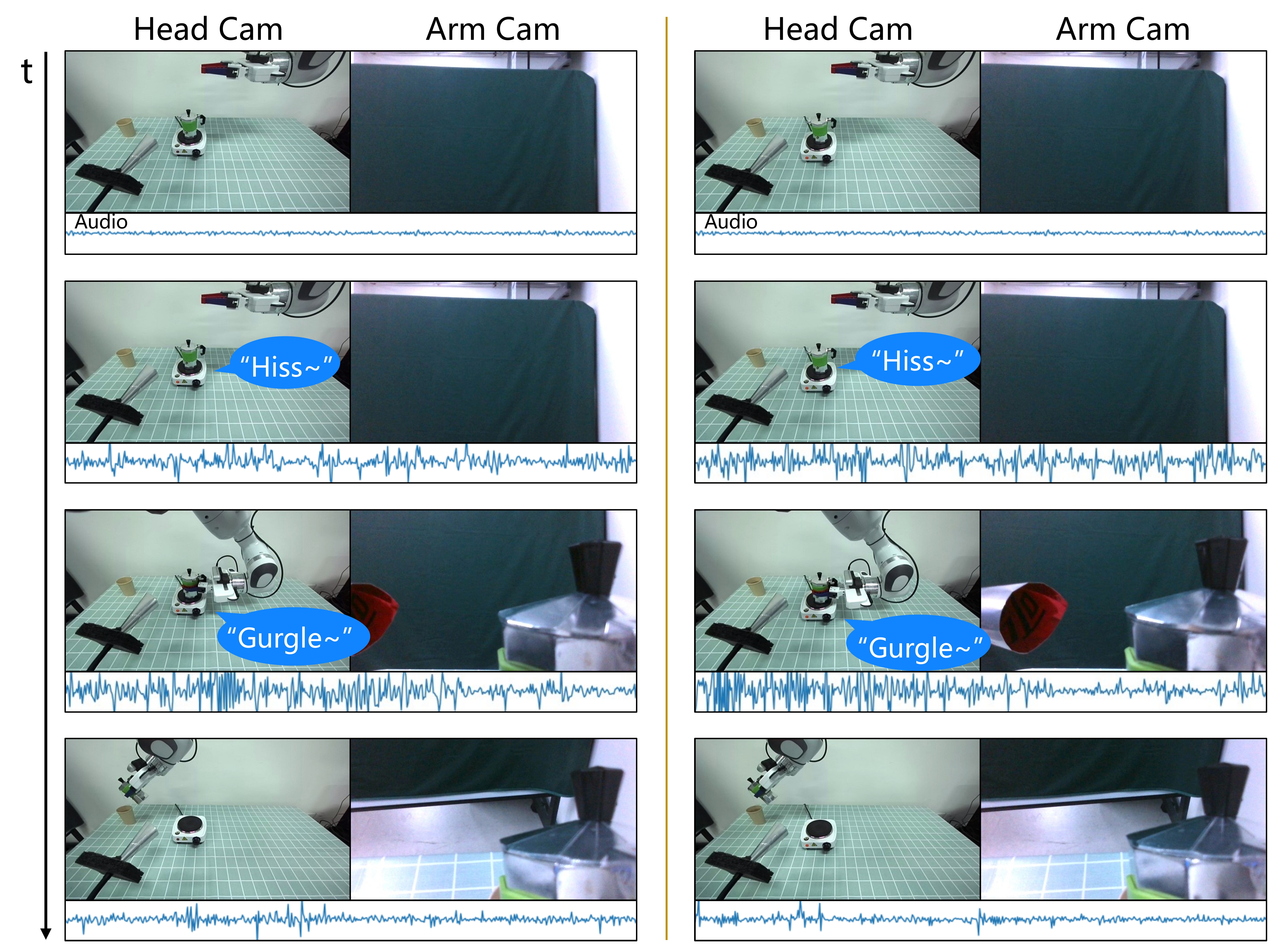}
\caption{\textbf{Real-robot physical deployment: \task{Moka Coffee}.} The robot monitors the complex acoustics of an uninstrumented Moka pot and initiates pouring only after detecting the characteristic sputtering phase. The vertical axis of the audio waveform is scaled by 10x for visualization.}

\label{fig:real_moka_coffee_scene}
\end{figure}

\paragraph{\task{Moka Coffee}.}
For the \task{Moka Coffee} evaluation, the physical robot is stationed near a Moka pot and a serving cup, as shown in Figure~\ref{fig:real_moka_coffee_scene}. The brewing process proceeds for an extended period, during which the visual scene remains largely static. Because a traditional Moka pot lacks electronic sensors or discrete visual indicators of completion, the robot cannot rely on digital feedback or simple visual heuristics. Instead, the robot must remain still and monitor the evolving sound of the boiling water and steam. It must grasp the pot and pour the coffee strictly after the acoustic profile transitions into the characteristic sputtering phase, which indicates that the extraction is complete. Pouring early is counted as a failure.

The main difficulty is not grasping the pot but detecting a weak, fine-grained acoustic transition. Under the earlier microphone setup, HEAR reached 0.18 success, compared with 0.05 for $\pi_{0.5}$-Waveform and 0.05/0.04 for ManiWAV and Play it by Ear. When the microphone is placed closer to the pot and the input gain is calibrated for the brewing sound, HEAR improves to 0.51. This result suggests that causal memory helps track the process, but reliable deployment still depends on capturing a sufficiently clear acoustic signal.

\begin{figure}[t]
\centering
\includegraphics[width=\linewidth]{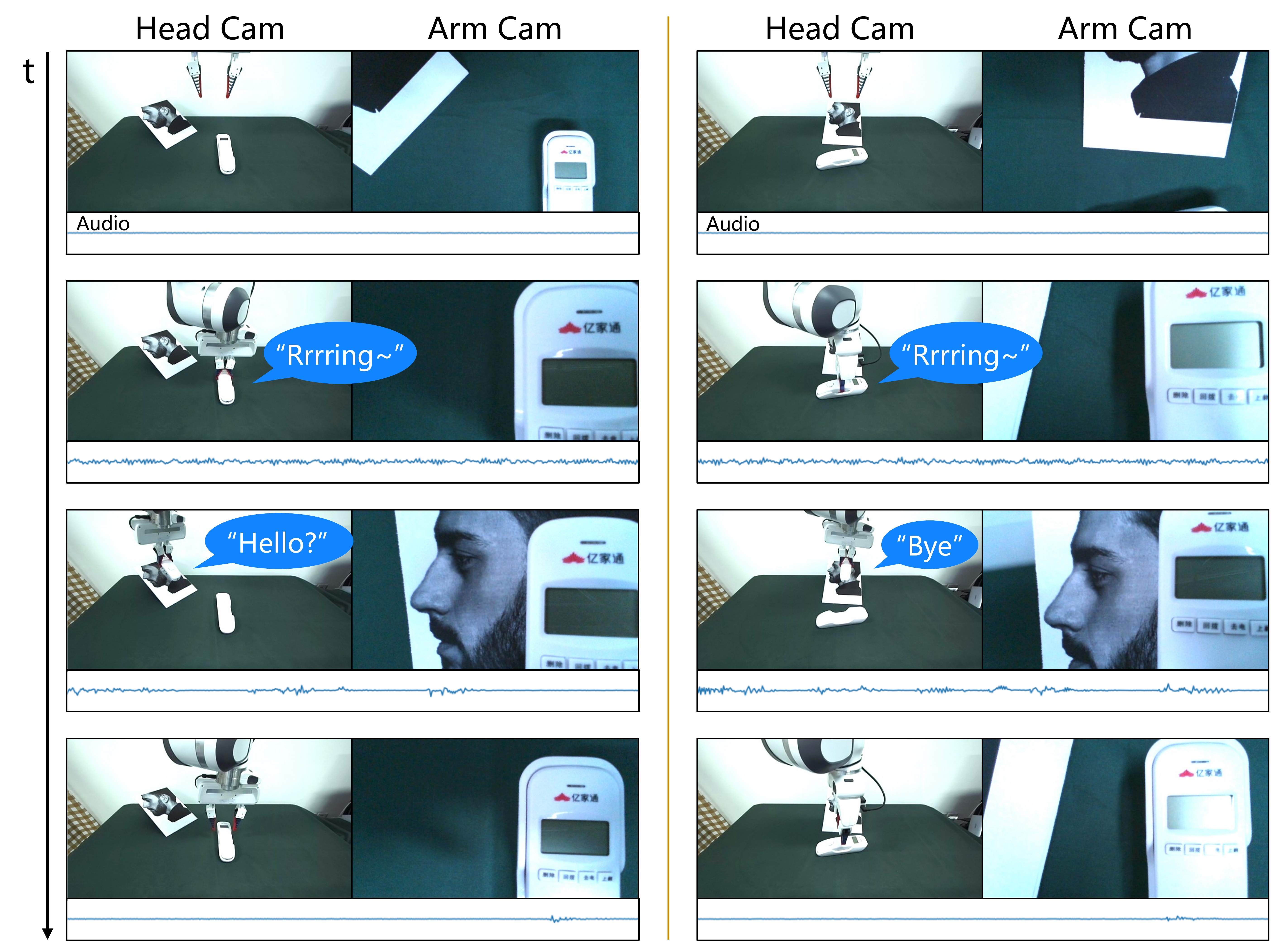}
\caption{\textbf{Real-robot physical deployment: \task{Answer Phone}.} Despite severe visual aliasing, the robot successfully waits for a ringtone, hands the phone to a user, and returns it to the base only after the acoustic end-of-call signal.}

\label{fig:real_answer_phone_scene}
\end{figure}

\paragraph{\task{Answer Phone}.}
As shown in Figure~\ref{fig:real_answer_phone_scene}, the \task{Answer Phone} sequence begins with the robot waiting for a randomized ringtone, after which it must retrieve a handset from a box and hand it to a person. During the call, it listens to speech. When the call ends, indicated by a hang-up sound or end-of-call speech, it must return the phone to the box.

This long-horizon task combines multiple sound categories and exposes severe visual aliasing. From a single RGB frame, the robot near the phone box looks nearly identical when it should pick up the ringing phone and when it should return the phone after a call. Correct behavior therefore depends entirely on progress tracking and causal history rather than snapshot visual recognition.

Under the earlier setup, HEAR reached a 0.15 success rate, compared with 0.06 for $\pi_{0.5}$-Waveform and 0.05 for ManiWAV. The dominant challenge is not only the ringtone onset, but also judging dialogue content and the end-of-call moment under repeated visual states. Expanding the task-specific demonstrations with more dialogue variations raises HEAR to 0.43. This indicates that causal memory is necessary for preserving the acoustic progress state, while dialogue diversity remains important for robust semantic timing.

\begin{figure}[t]
\centering
\includegraphics[width=\linewidth]{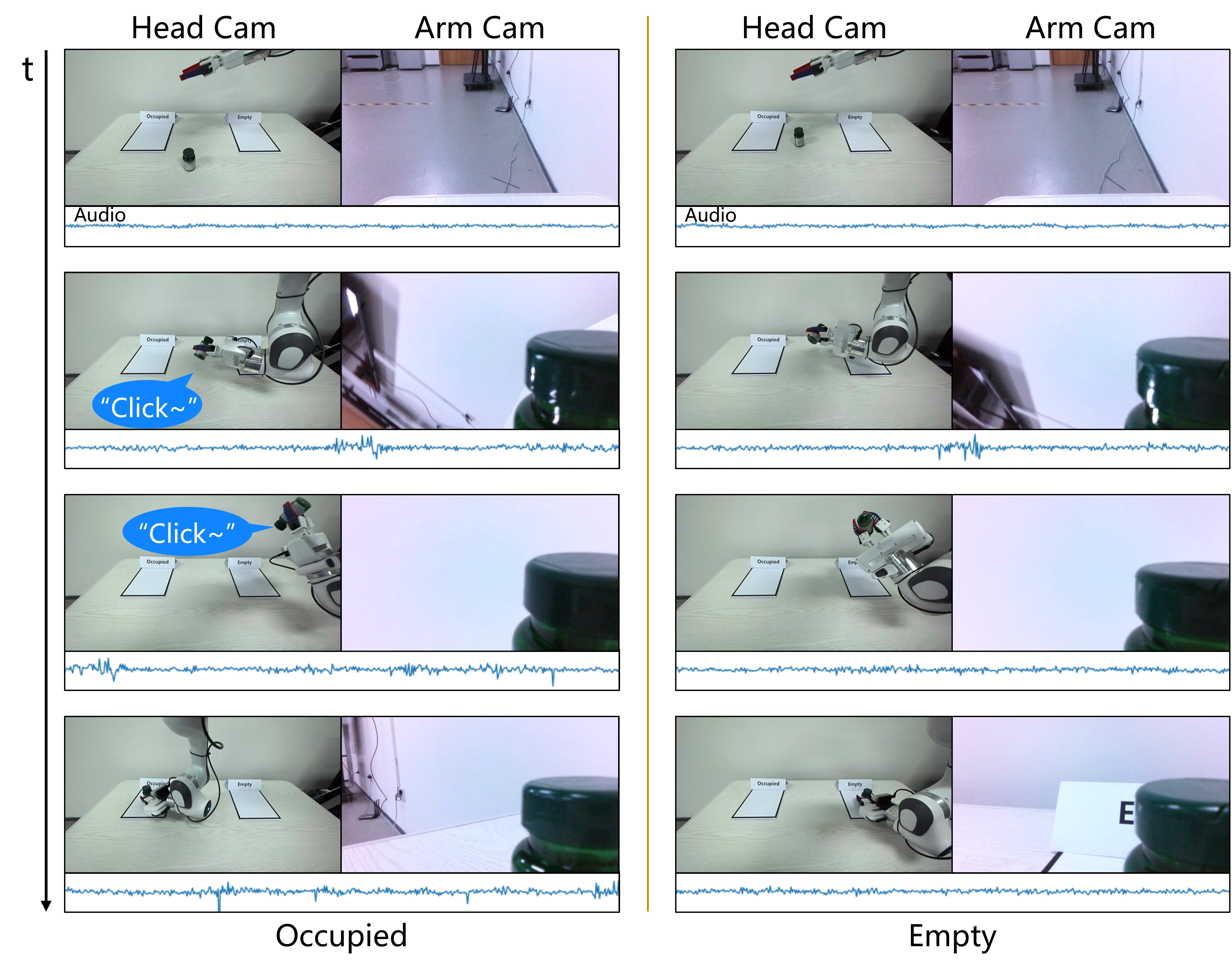}
\caption{\textbf{Real-robot physical deployment: \task{Shake Bottle}.} The robot performs a deliberate shaking motion to actively elicit sound from an opaque bottle, using the resulting rattle to classify its hidden contents. The vertical axis of the audio waveform is scaled by 10x for visualization.}
\label{fig:real_shake_bottle_scene}
\end{figure}

\paragraph{\task{Shake Bottle}.}
To test active acoustic sensing, \task{Shake Bottle} prompts the robot to grasp an opaque bottle and perform a brisk shaking motion to elicit sound. Based on the resulting rattle, the robot classifies the bottle as occupied or empty, as depicted in Figure~\ref{fig:real_shake_bottle_scene}. It then places the bottle into the corresponding region.

The bottle is randomly placed before each episode, and the policy begins only after the bottle is stationary and the workspace is silent. This ensures that placement noise cannot serve as a useful cue. The task is designed to reflect an unattended manipulation scenario involving an opaque bottle: because the robot cannot visually inspect the bottle's contents, it must actively interact with the bottle to generate sound and then use the self-generated rattling response to infer its internal state.

Unlike passive listening, this setup is a classic example of interactive perception. The acoustic evidence depends entirely on the robot's own motion. The same bottle can sound quiet if the shake is too weak. Success demands coupling audition with proprioception and generating repeatable motion primitives.

HEAR reaches a 0.88 success rate. Waveform rendering remains the strongest baseline ($\pi_{0.5}$-Waveform at 0.54), while vision-only variants are weak (0.22 to 0.48). HEAR achieves this result by fusing audio directly with joint positions (qpos) and producing smoother, highly repeatable shaking trajectories via the Realizer, ensuring the elicited acoustic evidence is consistent across trials.

\begin{figure}[t]
\centering
\includegraphics[width=\linewidth]{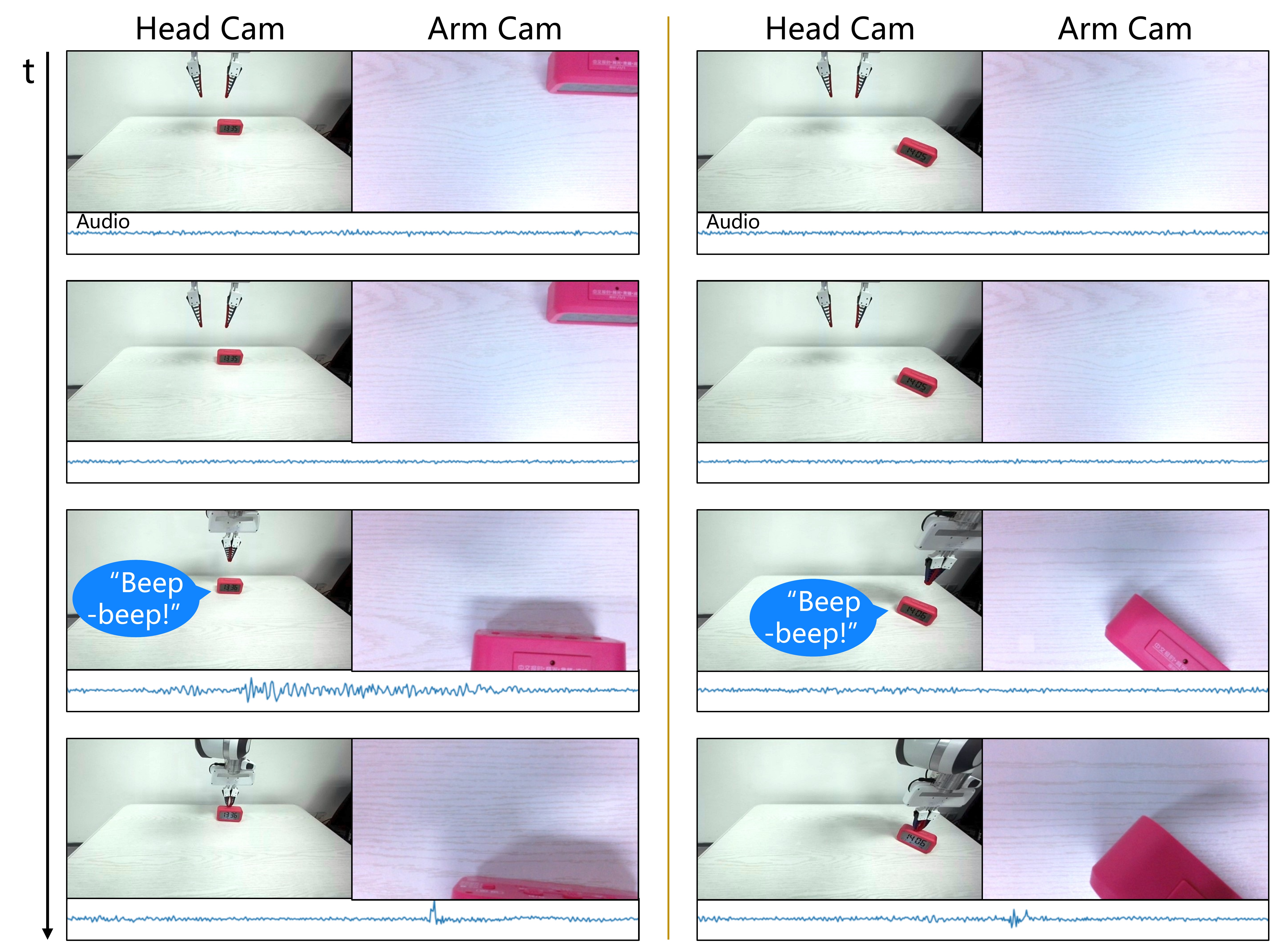}
\caption{\textbf{Real-robot physical deployment: \task{Real Alarm Clock}.} Operating under real-world background noise, the robot suppresses early visual heuristics and presses the button strictly after the sustained ring begins. The vertical axis of the audio waveform is scaled by 10x for visualization.}
\label{fig:real_real_alarmclock_scene}
\end{figure}

\paragraph{\task{Real Alarm Clock}.}
Deploying the alarm scenario in the physical world, \task{Real Alarm Clock} features a device that produces a sustained ring at an unpredictable moment. Figure~\ref{fig:real_real_alarmclock_scene} demonstrates how this setup requires the robot to suppress early actions and press the button only after the acoustic onset.

This setting tests whether policies remain sound-causal under physical domain shifts, including room reverberation and background chatter, and whether they can suppress the visually tempting early-press shortcut.

HEAR reaches a 0.96 success rate. Waveform rendering performs robustly ($\pi_{0.5}$-Waveform achieves 0.90), and ASR variants are competitive on this sustained cue ($\pi_{0.5}$-ASR at 0.78). Vision-only baselines fail frequently (0.01 to 0.04). In failed baseline trials, waveform images are occasionally destabilized by room reverberation which alters the visual envelope, leading to threshold artifacts. The text interface of ASR inherently compresses temporal detail, making reaction timing less precise. HEAR rarely fails on this task because it operates directly on waveform-derived embeddings and integrates evidence continuously, which suppresses premature presses and missed onsets in this acoustically salient setting.

\subsection{System Analysis and Ablations}
\label{sec:exp_ablation}
We conduct extensive ablations in the HEAR-Bench simulation environment to isolate the mechanisms driving performance and to understand the underlying challenges of sound-centric manipulation. To ensure a fair comparison, each ablation uses the same demonstrations, observation interface, and action representation, changing only the targeted component. Unless stated otherwise, we report the average task success rate over the seven simulation tasks based on 100 trials per task.

\begin{table}
\small\sf\centering
\caption{\textbf{Component ablations in HEAR-Bench.} We report the average success rate across all seven simulation tasks, demonstrating the contribution of each architectural module and training phase.}
\label{tab:ablations}
\begin{tabular}{l|c}
\toprule
Variant & Avg. success $\uparrow$ \\
\midrule
\textbf{HEAR full (OpenX-Sound pretrained)} & \textbf{0.81}\\
No OpenX-Sound pretraining & 0.69\\ \hline
No Historizer & 0.57\\
Historizer: causal GRU memory & 0.67\\
Historizer: EMA/pooling memory & 0.62\\ \hline
No Advancer & 0.73\\
No stage text & 0.77\\
No low-level controller & 0.75\\
Realizer regression & 0.70\\
\bottomrule
\end{tabular}
\end{table}

\subsubsection{Component Ablations}
\label{sec:exp_ablation_components}

Table~\ref{tab:ablations} isolates the contribution of individual architectural choices. First, removing the OpenX-Sound pretraining phase causes a substantial drop in average success (from 0.81 to 0.69). Because the base omni-modal backbone is already pretrained on broad general multimodal internet data, this gain is interpreted as evidence for robotics-specific audio-visual-action temporal alignment rather than generic multimodal scale.

Removing the Historizer results in the most severe performance decline (dropping to 0.57). This supports our core premise that a causal memory mechanism is necessary to preserve transient acoustic evidence across discrete decision boundaries. We also tested lighter memory alternatives, such as a causal GRU (0.67) and an exponential moving average (EMA) pooling over audio features (0.62). While these improve upon a memoryless design, they remain noticeably below our Streaming Stateful Transformer (SST). This suggests that the additional capacity and streaming attention provided by SST translate directly to better sound-causal control.

The Supplemental Material provides more detailed and comprehensive ablations and diagnostic analyses. Supplemental Section~S5.1 and Table~S1 report the Advancer training cost; Section~S5.2 and Table~S2 analyze temporal motion collapse; Section~S5.3 and Table~S3 test execution wrappers; Section~S5.4 and Figures~S1--S2 sweep audio-window length and chunk horizon; Section~S5.5 and Table~S4 decompose false-trigger and miss-detection errors; Section~S5.6 and Table~S5 provide real-world module ablations; Section~S5.7 and Table~S6 provide controlled OpenX-Sound pretraining variants; Section~S5.8 and Table~S7 report robustness sweeps; Section~S5.9 and Table~S8 decompose real-world failures; Section~S5.10 and Table~S9 analyze Advancer prediction quality; and Section~S5.11 and Table~S10 report deployment timing. These supplementary results support the same conclusion while keeping the main text centered on the primary simulation and physical experiments.

\section{Discussion and Future Work}
\label{sec:discussion}
The introduction of the Vision-Sound-Language-Action (VSLA) paradigm provides a useful perspective on multi-sensory robot learning. While modern vision-language-action models have achieved remarkable success by relying on persistent visual states, our findings suggest that closed-loop physical interaction often requires robots to process transient environmental signals in real time. In this section, we reflect on the broader implications of our framework, discuss data and deployment challenges, and close with limitations and future directions.

\subsection{Broader Implications for Multisensory Control}
The core structural challenge addressed by HEAR extends beyond just audio processing. It reflects a universal tension in modern robot systems between high-frequency, transient sensory events and low-frequency, delayed policy updates. In typical manipulation environments, visual states degrade gracefully under system delay because objects generally remain in the camera's view. Acoustic events, however, are highly ephemeral. A brief impact sound or a completion beep can easily occur within a single open-loop action chunk and then reach a memoryless policy only as missing, low-duty-cycle, or timing-ambiguous evidence.

Our work shows that appending new sensory modalities to existing VLA architectures is insufficient when the decisive evidence is transient. When dealing with transient signals under inevitable end-to-end system latency, closed-loop control requires deliberate architectural design. Specifically, it necessitates decoupling the high-frequency sensing cadence from the lower-frequency decision cadence. By using a continuous causal memory to bridge execution gaps and predictive dynamics to ground the policy in time, HEAR reduces critical information loss. This system-level principle is relevant not only for auditory perception but also for integrating other high-frequency physical modalities into large-scale robot models. For instance, recent efforts to unlock physical knowledge through tactile sensing face similar challenges regarding the transient nature of contact forces \citep{huang2025tactile}. Architectures that maintain causal persistence across execution gaps are a practical prerequisite for using these sensory inputs in closed-loop control.

\subsection{The Data Bottleneck in Audio-Driven Robotics}
Scaling the VSLA paradigm naturally encounters a severe lack of synchronized, real-world audio data in existing robot learning datasets. While our OpenX-Sound dataset provides a scalable workaround by synthesizing temporally aligned audio from video, we must explicitly acknowledge the domain gap this introduces. Advanced video-to-audio generation models can produce perceptually plausible soundscapes, but they do not perfectly capture true physical acoustics. Generated audio may occasionally omit subtle contact cues, alter precise event timings, or hallucinate sounds that lack a true physical source.

Because sound-centric tasks are highly sensitive to causality and timing, these synthetic artifacts limit the ceiling of purely offline pretraining. Consequently, OpenX-Sound serves primarily as a bootstrapping mechanism for representation learning rather than a complete substitute for physical data. This motivates native datasets that include synchronized microphone recordings alongside standard metadata, such as microphone placement and room acoustic properties. Fine-tuning on real-world, task-specific demonstrations remains a necessary step for reliable physical deployment until such datasets become widely available.

\subsection{Long-Horizon Multi-Stage Interaction}
\label{sec:limit_long_horizon}
The long-horizon physical tasks remain deployment-sensitive settings. Under the uniform acoustic setup, \task{Moka Coffee} and \task{Answer Phone} are much harder than shorter tasks with more salient cues; task-calibrated microphone placement, gain adjustment, and richer dialogue variation improve them to 51\% and 43\%, respectively. The supplementary failure decomposition explains the dominant sources of difficulty: \task{Moka Coffee} depends on a low-energy, fine-grained sputtering transition, whereas \task{Answer Phone} depends on sound-type discrimination and end-of-dialogue timing. These bottlenecks reflect a broader system-level challenge for long-horizon embodied policies: multi-stage tasks are difficult not only for VSLA but also for VLA systems, because they require persistent state, recovery from earlier uncertainty, and stable decision making across repeated observations. More robust deployment therefore requires system-level solutions, including better microphone geometry, active source separation, stage-level recovery policies, and longer-term task memory in addition to causal acoustic memory.

The deployed BEI is also not eliminated by faster inference alone. Software optimization and shorter action chunks can reduce the gap, but transient sounds can still occur between policy queries, and the supplementary execution-wrapper analysis shows that overly aggressive replanning can introduce discontinuous motion and ego-noise. This is why the Historizer is treated as a structural component rather than a substitute for engineering optimization, and why longer-term memory remains a worthwhile future direction for both VLA and VSLA systems.

\subsection{Synthetic Audio Domain Gap}
\label{sec:limit_synth}
OpenX-Sound is designed as a pretraining resource, not as a replacement for native robot audio. Synthetic tracks are useful because they provide broad temporal co-occurrence between video, action, and plausible sound, but they cannot fully reproduce room acoustics, microphone placement, ego-noise, or the fine attack profile of contact events. This gap is most visible in the real-robot results: pretraining improves representation learning, yet all physical deployment numbers still require task-specific fine-tuning with recorded robot audio. We therefore treat OpenX-Sound as a scalable bootstrap for VSLA learning, while native synchronized audio remains necessary for final deployment.

\subsection{Limitations and Future Directions}
While our approach provides a concrete step toward sound-centric manipulation, the current system has several limitations.

First, our hardware setup relies on a single workspace microphone. This configuration inherently limits the robot's ability to localize sound sources or reliably separate overlapping acoustic events in noisy environments, a challenge functionally similar to the classic cocktail party problem \citep{cherry1953some}. Spatial hearing capabilities, such as combining a base microphone with an end-effector microphone, provide spatial acoustic cues and improve signal-to-noise ratios through active beamforming.

Second, our policies are trained entirely via offline imitation learning. Consequently, deployment robustness depends on the diversity of acoustic noise and timing variations present in the demonstration data. Real-world deployment introduces mechanical ego-noise and room reverberation that can degrade perception. Online fine-tuning and reinforcement learning methods that adapt the robot to its specific acoustic environment on the hardware remain promising ways to improve deployment reliability~\citep{lu2025vla}.

Finally, our current observation interface processes audio and vision as distinct streams before fusing them at the representation level. In human perception, however, subtle visual micro-motions are often deeply synchronized with sound generation. Native audio-visual video encoders that process synchronized multimodal clips directly are a promising direction~\citep{kim2026cosmos}. This integration can help robots better associate specific acoustic timbres with physical contact events, further reducing visual aliasing and paving the way for more holistic physical intelligence.

\section{Conclusion}
\label{sec:conclusion}
This paper formalized the Vision-Sound-Language-Action (VSLA) paradigm for sound-dependent manipulation, where transient acoustic cues must be preserved despite delayed, chunked policy execution. We proposed HEAR, which decouples high-frequency auditory memory from low-frequency action generation through a streaming Historizer, temporal Advancer, multimodal Envisioner, and flow-matching Realizer.

We introduced OpenX-Sound and HEAR-Bench to support pretraining and causally strict evaluation, and validated HEAR in both simulation and real-robot deployments. The results show that audio-reactive manipulation requires more than appending an audio input to existing VLA models: it needs causal auditory persistence and explicit temporal grounding. At the same time, long-horizon physical tasks such as coffee monitoring and phone interaction remain challenging, leaving robust acoustic sensing, source separation, and stage-level recovery as important future directions.

\begin{orcidids}
Chang Nie \url{https://orcid.org/0009-0002-7867-3363}\\
Tianchen Deng \url{https://orcid.org/0000-0002-4368-7936}\\
Hesheng Wang \url{https://orcid.org/0000-0002-9959-1634}
\end{orcidids}

\begin{funding}
The authors disclosed receipt of the following financial support for the research, authorship, and/or publication of this article: This work was supported by the National Key R\&D Program of China under Grants 2024YFB4708900, in part by the National Natural Science Foundation of China under Grants 62225309, U24A20278, and 62361166632, in part by the Explorers Program of Shanghai (Basic Research Funding) under Grant 25TS1415100, and in part by the Shanghai Municipal Special Program for Basic Research on General AI Foundation Models under Grant 2025SHZDZX025G01.
\end{funding}

\begin{dci}
The authors declared no potential conflicts of interest with respect to the research, authorship, and/or publication of this article.
\end{dci}

\begin{sm}
Supplemental material for this article is available online.
\end{sm}

\bibliographystyle{SageH}
\bibliography{hear_refs}

\end{document}